%% file: main.tex
\documentclass{article}

\PassOptionsToPackage{numbers, compress}{natbib}

     \usepackage[preprint]{neurips_2021}

\usepackage[utf8]{inputenc} 
\usepackage[T1]{fontenc}    
\usepackage{hyperref}       
\usepackage{url}            
\usepackage{booktabs}       
\usepackage{amsfonts}       
\usepackage{nicefrac}       
\usepackage{microtype}      
\usepackage{xcolor}         
\usepackage{graphicx}
\usepackage{subcaption}
\usepackage{bbm}
\usepackage{wrapfig}

\input{symbols}

\title{RAPTOR: End-to-end Risk-Aware MDP Planning and Policy Learning by Backpropagation}

\author {
    Noah Patton\thanks{Authors contributed equally.} \\
    University of Toronto \\
    \texttt{noah.patton@mail.utoronto.ca} \\
    \And Jihwan Jeong\footnotemark[1] \\
    University of Toronto \\
    \texttt{jhjeong@mie.utoronto.ca} \\
    \And Michael Gimelfarb\footnotemark[1]\text{ } \thanks{Affiliate to Vector Institute, Toronto, Canada.} \\
    University of Toronto \\
    \texttt{mike.gimelfarb@mail.utoronto.ca} \\
    \And
    Scott Sanner\footnotemark[2] \\
    University of Toronto \\
    \texttt{ssanner@mie.utoronto.ca}
}

\newcommand{\acronym}{RAPTOR}
\newcommand{\tfplan}{BackpropPlan}

\makeatletter
\newcommand{\appendixref}[2]{\@ifundefined{r@#1}{#2}{\ref{#1}}}
\makeatother

\newif\ifappendix
\appendixtrue 

\begin{document}

\maketitle

\begin{abstract}
    Planning provides a framework for optimizing sequential decisions in complex environments. Recent advances in efficient planning in deterministic or stochastic high-dimensional domains with continuous action spaces leverage backpropagation through a model of the environment to directly optimize actions. However, existing methods typically not take risk into account when optimizing in stochastic domains, which can be incorporated efficiently in MDPs by optimizing the entropic utility of returns. We bridge this gap by introducing Risk-Aware Planning using PyTorch (\acronym), a novel framework for risk-sensitive planning through end-to-end optimization of the entropic utility objective. A key technical difficulty of our approach lies in that direct optimization of the entropic utility by backpropagation is impossible due to the presence of environment stochasticity. The novelty of \acronym~lies in the reparameterization of the state distribution, which makes it possible to apply stochastic backpropagatation through sufficient statistics of the entropic utility computed from forward-sampled trajectories. The direct optimization of this empirical objective in an end-to-end manner is called the risk-averse straight-line plan, which commits to a sequence of actions in advance and can be sub-optimal in highly stochastic domains. We address this shortcoming by optimizing for risk-aware Deep Reactive Policies (RaDRP) in our framework. We evaluate and compare these two forms of \acronym~on three highly stochastic domains, including nonlinear navigation, HVAC control, and linear reservoir control, demonstrating the ability to manage risk in complex MDPs.
\end{abstract}

\section{Introduction}
\label{sec:intro}

As machine learning models are more frequently deployed in the real world, the concern over ensuring their safety has been ever-increasing \citep{faria2018,pereira2020}. In sequential stochastic decision-making problems, it has been shown that optimizing the expected cumulative reward can lead to undesirable outcomes such as excessive risk-taking, since low-probability catastrophic outcomes with negative reward, or \emph{risk}, can be underrepresented \citep{moldovan2014safety}. The \emph{risk-averse MDP} framework addresses this problem by optimizing risk measures with favorable mathematical properties \citep{ruszczynski2010risk}. 

On the other hand, \emph{planning} optimizes decisions or actions given a mathematical description of the environment, thus minimizing the need to do dangerous exploration in the real world. However, despite advances in scalable end-to-end planning, existing approaches do not typically take risk into account. For instance, \emph{\tfplan}\ \citep{wu2017scalable} utilizes recent advances in deep learning and is highly scalable in continuous state or action spaces (CSA-MDPs). In this framework, the transition model and reward are encoded in RNN-like cells. Unlike a typical neural network, the inputs to the network are the actions that are optimized through backpropagation. By employing highly effective tools for non-convex optimization \citep{rmsprop}, \tfplan\ is able to efficiently learn optimal sequences of actions. However, its main limitation is that it cannot be applied to stochastic models, which could be addressed by learning \emph{reactive policies} \citep{bueno2019deep}. However, neither approach incorporates risk explicitly.

One popular approach for incorporating risk-sensitivity into MDPs is the entropic utility, as well as its mean-variance approximation, which can be interpreted as trade-offs between the mean and the variance of the cumulative return \citep{bauerle2014more}. One key advantage of the entropic utility is that it is defined entirely in terms of expectation operators, rather than percentiles or other quantities that can more difficult to estimate from sample trajectories, making it an intuitive starting point for scalable planning. Furthermore, the entropic utility satisfies a recursive property \citep{osogami2012robustness} that makes it possible to optimize directly in sequential decision-making problems without relying explicitly on the Bellman principle, which often presents computational challenges in other risk-aware MDP and reinforcement learning frameworks \citep{defourny2008risk,mannor2011mean}. This makes it particularly suitable for gradient-based end-to-end planning. 

\begin{figure}[!t]
    \centering
    \includegraphics[width=1.0\linewidth]{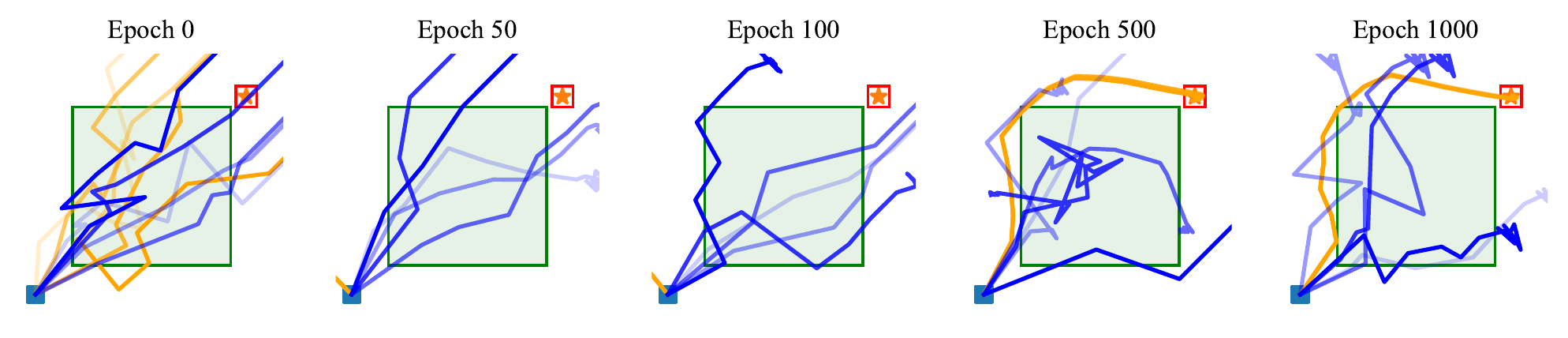}
    \caption{The evolution of trajectories navigated by \acronym\ (orange) and a risk-neutral agent (blue) in a two-dimensional Navigation domain subject to stochastic dynamics (note: multiple sample trajectories are shown simultaneously in each epoch). The green bounding box at the center is a high-variance zone. The more an agent traverses into the box the higher the variability in the next position at which the agent lands. The blue square on the bottom left is the starting position, while the red box at the upper right corner shows the goal region. 
    By epoch $1000$, we clearly see that the risk-sensitive agent is able to get to the goal region by avoiding the box and thus failure, while the risk-neutral agent does not.
    }
    \label{fig:new_nav_sidebyside}
\end{figure}

\begin{wrapfigure}{r}{0.4\linewidth}
    \includegraphics[width=\linewidth]{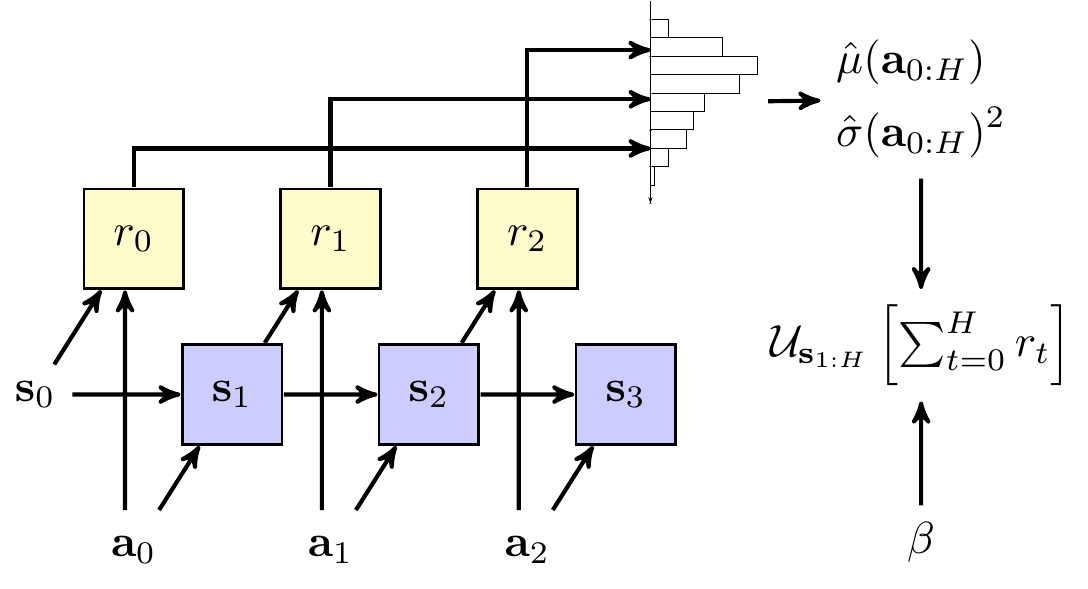}
    \caption{A simplified decision diagram for planning with an entropic utility objective of the return, $\utility{}{\sum_t r_t}$. Here, the sufficient statistics of the return distribution, namely mean $\hat{\mu}$ and variance $\hat{\sigma}^2$ are computed by forward sampling state trajectories and used to approximate the utility. Applying the reparameterization trick and backpropagation will lead to our overall approach \acronym.}
    \label{fig:workflow_simple}
\end{wrapfigure} 
To this end, we propose \acronym~(\emph{Risk-Aware Planning using pyTORch}), which enables scalable risk-aware end-to-end planning for CSA-MDPs by leveraging automatic differentiation~\citep{pytorch} for gradient-based optimization w.r.t. the MDP model and an entropic utility objective. To achieve this, we begin by leveraging an extension of \tfplan\ to accommodate stochastic transitions \citep{bueno2019deep}, by representing the planning domain as a stochastic computation graph (Section \ref{ssec:stochastic-graph}). While we cannot directly formalize the distributionally-defined entropic utility in closed-form for end-to-end planning, 
for many classes of problems we can reparameterize the objective and apply stochastic backpropagation. This allows us to compute the entropic utility directly from the sufficient statistics of forward-sampled trajectories, while still permitting backpropagation through the utility and its symbolic computation from sufficient statistics of trajectories. We then leverage these computational results to propose two approaches for optimizing \emph{risk-aware} decisions (Section \ref{ssec:drp}). The first derives a risk-sensitive straight-line plan, in which the agent commits to a sequence of actions in advance, while the second learns a \emph{deep reactive policy} and is particularly suited for highly stochastic domains. Both approaches integrate seamlessly into our optimization framework, and can be computed efficiently in an end-to-end manner by reparameterization and backpropagation. 

Indeed, as evidenced by Figure \ref{fig:new_nav_sidebyside}, \acronym~is effective at finding risk-averse behaviors in highly stochastic environments in a computationally efficient manner. Overall, empirical evaluations in Section \ref{sec:experiments} on three highly stochastic domains involving continuous action parameters --- navigation, HVAC control, and reservoir control --- demonstrate that \acronym\ is a reliable and efficient end-to-end method for risk-sensitive planning in complex MDPs.  

\section{Preliminaries}
\label{sec:prelim}

\subsection{Continuous State-Action Markov Decision Process (CSA-MDP)}
\label{ssec:mdp}

Sequential decision-making problems in this work are modeled as \emph{continuous state-action Markov decision processes} (CSA-MDPs), defined as tuples $\langle \states, \actions, \rewardfunction, \dynamicsfunction, \state{0}\rangle$: $\states\subseteq \mathbb{R}^{n}$ is the state space, $\actions\subseteq\mathbb{R}^m$ is the action space, $\rewardfunction : \states \times \actions \to \reals$ is a bounded differentiable reward function, $\dynamicsfunction : \states \times \actions \times \states \to [0, \infty)$ describes the non-linear dynamics of the system, and $\state{0}$ is the initial state. Note that CSA-MDPs are naturally factored \citep{factored-mdp-boutilier}, such that the state components are mutually independent given the previous state $\state{t}$ and action $\action{t}$. 

In the \emph{risk-neutral} setting, the objective function to optimize is the expected \emph{return},
\begin{equation}
\label{eqn:risk_neutral_value}
    V_h(\action{h:H}) \defeq \E{\state{h+1:H}}{\sum_{t=h}^{H} \reward{\state{t}}{\action{t}}},
\end{equation}
where state trajectories $\state{h+1:H} = \state{h+1},\state{h+2},\dots \state{H}$ are sampled according to $\dynamicsfunction$ and actions can either be computed from a closed-loop \emph{reactive policy} $\pi$, e.g. $\action{t} = \pi_t(\state{t})$, or an open-loop policy or \emph{plan} $\action{h:H} = \action{h}, \action{h+1}, \dots \action{H}$. 

\subsection{Risk-Aversion in CSA-MDPs}
\label{ssec:risk}

Risk sensitivity can be incorporated into the agent's decision-making by replacing the expectation operator $\E{\state{h+1:H}}{\cdot}$ with a non-linear utility function. In this paper, we consider the \emph{entropic utility}, that for $\beta \in \reals$ and a random variable $X$ is defined as
\begin{equation}
\label{eqn:entropic}
    \utility{}{X} \defeq \frac{1}{\beta}\log \E{}{e^{\beta X}}.
\end{equation}
Taylor expansion of (\ref{eqn:entropic}) obtains the mean-variance approximation
\begin{equation}
\label{eqn:mean_variance}
    \utility{}{X} = \E{}{X} + \frac{\beta}{2} \var{}{X} + O(\beta^2).
\end{equation}
Now, interpreting variance as risk, $\beta$ can be interpreted as the overall level of risk aversion of the agent: $\beta = 0$ induces \emph{risk-neutral} behavior, while choosing $\beta > 0$ ($\beta < 0$) induces \emph{risk-seeking} (\emph{risk-averse}) behaviors. Thus for a risk-averse agent, the entropic utility can provide protection against high return variability. In addition, the entropic utility is perhaps the most well-known \emph{convex/concave utility}, which means it is monotone, translation invariant, and concave for $\beta < 0$, properties that are often seen as minimal requirements for developing rational decision-making \citep{follmer2002convex,maccheroni2006ambiguity}. 

Generalizing (\ref{eqn:risk_neutral_value}) to the risk-aware setting, we define the \emph{utility} of an action sequence as
\begin{equation}
\label{eqn:utility_value}
    U_h(\action{h:H}) \defeq \utility{\state{h+1:H}}{\sum_{t=h}^{H} \reward{\state{t}}{\action{t}}},
\end{equation}
where it is understood that expectations are computed w.r.t. the distribution of $\state{h+1:H}$. Furthermore, due to the \emph{recursive property} of entropic utility \citep{bauerle2014more,dowson2020multistage,osogami2012robustness}, the optimal utility-to-go $\utilityopt{h}{\state{h}} \defeq \sup_{\pi}\, U_h(\action{h:H})$ over closed-loop policies satisfies the \emph{Bellman equation}
\begin{equation}
\label{eqn:utility_bellman_equation}
    \utilityopt{h}{\state{h}} = \sup_{\action{h} \in \actions}\,\utility{\state{h+1}}{\reward{\state{h}}{\action{h}} + 
    \utilityopt{h+1}{\state{h+1}}},
\end{equation}
that is analogous to the risk-neutral setting \citep{puterman2014markov}. Furthermore, the entropic utility is the \emph{only} convex/concave utility that can be written in this way \citep{kupper2009representation}. This makes the entropic utility particularly suitable for learning policies and planning in CSA-MDPs.

\section{Risk-Aware Planning}
\label{sec:main}

The goal of planning is to avoid the expensive computation of \eqref{eqn:utility_bellman_equation}. In this section, we formally define our proposed method, \acronym, that computes a policy or plan for the risk-sensitive objective \eqref{eqn:utility_value} by optimizing it directly in an end-to-end manner. To this end, we employ stochastic computation graphs and the reparameterization trick for the entropic utility.

\subsection{Risk-Aware End-to-End Planning by Backpropagation}
\label{ssec:stochastic-graph}

We begin with the idea of learning optimal open-loop policies for the utility objective (\ref{eqn:utility_value}) using gradient descent:
\begin{align}
    \pderiv{U_h(\action{h:H})}{\action{t}} 
    &= \pderiv{}{\action{t}}\left( \frac{1}{\beta} \log \E{\state{h+1:H}}{\exp{\left(\beta \sum_{t=h}^H \reward{\state{t}}{\action{t}}\right)}}\right) \nonumber \\
    \label{eqn:bad_gradient}
    &\propto \pderiv{}{\action{t}}\E{\state{h+1:H}}{\exp{\left(\beta \sum_{t=h}^H \reward{\state{t}}{\action{t}}\right)}},
\end{align}
where we have ignored the denominator for ease of exposition. However, backpropagating through the expectation (\ref{eqn:bad_gradient}) requires backpropagating through the entire sequence of state trajectories $\state{h},\state{h+1} \dots$, since $\state{t+1}\sim \dynamics{\state{t}}{\action{t}}{\cdot}$ depends implicitly on the actions. The difficulty therefore lies in our inability to directly backpropagate through the stochastic transitions that are not inherently differentiable. 

\paragraph{Reparameterization Trick}
Instead, we employ the widely-used \emph{reparameterization trick} \citep{blundell-icml-2015, implicit-reparameterization,vae-kingma, schulman-stochastic-gradient}, by rewriting a sample from the distribution $\state{t+1} \sim \dynamics{\state{t}}{\action{t}}{\cdot}$ as the output of a deterministic differentiable function $\phi : \states \times \actions \times \noises  \to \states$: 
\begin{equation}
\label{eqn:reparam}
    \state{t+1} = \phi(\state{t}, \action{t}, \noise{t}), \quad \noise{0}, \noise{1}, \dots \noise{H} \sim \noisedensity, \quad t = h, h + 1, \dots H,
\end{equation}
where $\noise{0:H} = \noise{0}, \noise{1}, \dots \noise{H}$ is a sequence of i.i.d. disturbances drawn from some distribution $\noisedensity$ on $\noises$, which we call a \emph{scenario}. For practical illustration, if $\dynamics{\state{t}}{\action{t}}{\cdot}$ belongs to a location-scale family, then $\phi(\state{t}, \action{t}, \noise{t}) = m(\state{t},\action{t}) + \noise{t} v(\state{t}, \action{t})$ for some functions $m$ and $v$, and can be generalized to other distributions \citep{ruiz2016generalized}. The reparameterization trick (\ref{eqn:reparam}) now allows us to rewrite the state trajectory explicitly as a function of the initial state $\state{h}$, previous actions and previous disturbances\footnote{For $t = h$ we simply define $\Phi_h(\state{h}, \action{h:t-1}, \noise{h:t-1}) \defeq \state{h}$ to be consistent with the derivations that follow. We also do the same for $\Psi_h$ defined in the following section.}:
\begin{align}
    \state{t} 
    &= \phi(\state{t-1}, \action{t-1}, \noise{t-1})\nonumber \\
    &= \phi(\phi(\state{t-2}, \action{t-2}, \noise{t-2}), \action{t-1}, \noise{t-1}) \nonumber \\
    \label{eqn:full_reparam_to_h}
    &= \dots \defeq \Phi_t(\state{h}, \action{h:t-1}, \noise{h:t-1}), \quad \forall t = h, h + 1, \dots H.
\end{align}
Plugging (\ref{eqn:full_reparam_to_h}) back into (\ref{eqn:bad_gradient}), the gradient can now be computed \emph{deterministically} by sampling a scenario $\noise{h:H} = \noise{h}, \noise{h+1}, \dots \noise{H}$ \emph{a priori}, and backpropagating through the expectation:
\begin{align*}
    \pderiv{U_h(\action{h:H})}{\action{t}} 
    &\propto \pderiv{}{\action{t}}\E{\state{h+1:H}}{\exp{\left(\beta \sum_{t=h}^H \reward{\state{t}}{\action{t}}\right)}} \\
    &= \pderiv{}{\action{t}}\E{\noise{h:H}}{\exp{\left(\beta \sum_{t=h}^H \reward{\Phi_t(\state{h}, \action{h:t-1}, \noise{h:t-1})}{\action{t}}\right)}} \\
    &= \E{\noise{h:H}}{\pderiv{}{\action{t}}\exp{\left(\beta \sum_{t=h}^H \reward{\Phi_t(\state{h}, \action{h:t-1}, \noise{h:t-1})}{\action{t}}\right)}}.
\end{align*}

\paragraph{Planning Through Sufficient Statistics of the Return Distribution}
While optimizing the exact entropic utility is possible, it can often overflow due to computing an exponential function of the return \citep{gosavi2014beyond}. An alternative approach is to approximate the utility by the mean-variance approximation (\ref{eqn:mean_variance}), in which the mean and variance of the return can be represented symbolically, and backpropagating through these sufficient statistics by again making use of the reparameterization trick (\ref{eqn:full_reparam_to_h}):
\begin{align}
    &U_h(\action{h:H})
    = \utility{\state{h+1:H}}{\sum_{t=h}^{H} \reward{\state{t}}{\action{t}}} \nonumber \\
    &\approx \E{\state{h+1:H}}{\sum_{t=h}^{H} \reward{\state{t}}{\action{t}}} + \frac{\beta}{2} \var{\state{h+1:H}}{\sum_{t=h}^{H} \reward{\state{t}}{\action{t}}} \nonumber \\
    \label{eqn:symbolic_mv}
    &= \E{\noise{h+1:H}}{\sum_{t=h}^{H} \reward{\Phi_t(\state{h}, \action{h:t-1}, \noise{h:t-1})}{\action{t}}} + \frac{\beta}{2} \var{\noise{h+1:H}}{\sum_{t=h}^{H} \reward{\Phi_t(\state{h}, \action{h:t-1}, \noise{h:t-1})}{\action{t}}}.
\end{align}
Now (\ref{eqn:symbolic_mv}) can be estimated via \emph{sample approximations}, in which a finite number $m$ of scenarios $\noise{h:H}^i \sim \prod_{t=h}^H \noisedensity(\noise{t})$ are forward-sampled to estimate $\state{h}^i, \state{h+1}^i, \dots \state{H}^i$ and $r_t^i \defeq \reward{\state{t}^i}{\action{t}}$:
\begin{equation}
\label{eqn:sample_approx}
    \begin{aligned}
        {U}_h(\action{h:H}) 
        &\approx \hat{U}_h(\action{h:H}) \defeq \hat{\mu}_h(\action{h:H}) + \frac{\beta}{2} \hat{\sigma}_h(\action{h:H})^2 \\
        \hat{\mu}_h(\action{h:H}) &\defeq \frac{1}{m} \sum_{i=1}^m \sum_{t=h}^H r_t^i, \quad 
        \hat{\sigma}_h(\action{h:H})^2 \defeq \frac{1}{m} \sum_{i=1}^m \left(\sum_{t=h}^H r_t^i - \hat{\mu}_h(\action{h:H}) \right)^2.
    \end{aligned}
\end{equation}
In other words, computing an optimal sequence of actions requires backpropagating through the sufficient statistics $\hat{\mu}_h(\action{h:H})$ and $\hat{\sigma}_h(\action{h:H})^2$ w.r.t. the action sequences $\action{h:H}$. Through reparameterization, the gradients of the sample sufficient statistics (\ref{eqn:sample_approx}) can now flow through $\phi$ end-to-end:
\begin{align*}
    \pderiv{\hat{U}_h(\action{h:H})}{\action{t}} 
    &\approx \pderiv{\hat{U}_h(\action{h:H}) }{\state{t+1}} \pderiv{\state{t+1}}{\action{t}} \\
    &= \pderiv{\state{t+1}}{\action{t}} 
        \sum_{\tau=t+1}^H \pderiv{\hat{U}_h(\action{h:H})}{r_\tau} 
        \pderiv{r_\tau}{\state{\tau}} \prod_{\kappa=\tau}^{t+2}\pderiv{\state{\kappa}}{\state{\kappa-1}} \\
    &= \pderiv{\phi(\state{t},\action{t},\noise{t})}{\action{t}} 
        \sum_{\tau=t+1}^H \pderiv{\hat{U}_h(\action{h:H})}{r_\tau} 
        \pderiv{r_\tau}{\state{\tau}} \prod_{\kappa=\tau}^{t+2} \pderiv{\phi(\state{\kappa-1},\action{\kappa-1},\noise{\kappa-1})}{\state{\kappa-1}},
\end{align*}
where we have ignored the existence of multiple rollouts with index $i$ for ease of notation. In essence, actions that optimize the utility can now be computed by forward-sampling $\noise{h:H}^i$, backpropagating through the utility's sufficient statistics computed on the simulated scenarios $\noise{h:H}^i$ and updating the actions by gradient descent, e.g. $\action{t}' = \action{t} - \alpha \pderiv{\hat{U}_h(\action{h:H})}{\action{t}}$.

\subsection{Risk-Aware Straight-Line Plan and Deep Reactive Policy}
\label{ssec:drp}

\paragraph{Risk-Aware Straight-Line Plan}
The optimization of $\hat{U}_h(\state{h})$ by backpropagation avoids the computationally expensive iteration of (\ref{eqn:utility_bellman_equation}) while providing a maximizing plan $\action{0:H}^*$. However, the plan computed above commits to a fixed sequence of actions starting from the initial state $\state{0}$ and follows them until the end of the planning horizon. For this reason, we call this plan the \emph{risk-aware straight-line plan} (SLP), and its corresponding utility
\begin{equation}
\label{eqn:straight-line}
    u_{SL}(\state{0}) \defeq \sup_{\action{0:H}}\, \utility{\noise{0:H}}{\sum_{t=0}^{H} \reward{\state{t}}{\action{t}}}.
\end{equation}
The limitation of an SLP is that its utility could be lower than the closed loop policy in (\ref{eqn:utility_bellman_equation}), since stochasticity prevents it from responding to significant deviations in states from their anticipated trajectory. On the other hand, (\ref{eqn:straight-line}) provides a (potentially tight) lower bound to the utility of the optimal closed-loop policy \citep{mercier2008amsaa}, and has been used extensively in risk-neutral planning \citep{burns2012anticipatory,issakkimuthu2015hindsight,raghavan2017hindsight}. We provide this lower bound in the risk-averse setting (see Appendix \appendixref{subsec:proof}{A.1}) for a proof).
\begin{theorem}
\label{thm:slp}
    The quantity $u_{SL}(\state{0})$ is a lower bound to the optimal utility, $\utilityopt{0}{\state{0}} \geq u_{SL}(\state{0})$.
\end{theorem}

\paragraph{Risk-Aware Deep Reactive Policy}
In order to optimize CSA-MDPs without facing the potential limitations of computing straight-line plans, an alternative approach is to learn a risk-aware \emph{reactive policy}, which is an explicit mapping from states to actions, e.g. $\pi_t : \states \to \actions$ \citep{bueno2019deep}. In order to realize efficient planning end-to-end, we can learn a parameterized policy $\pi_\theta$, where $\theta$ represent the weights of a parameteric model such as a neural network, in which case this reduces to a risk-aware instantiation of the \emph{deep reactive policy} (DRP).

\paragraph{Reparameterization Trick for DRPs}
The deep reactive policy depends on the state, which in turn depends implicitly on the previous action, which depends on the previous state, and so on, which makes the reparameterization of the entropic utility more involved. In this case, the reparameterization trick described in the previous section could be applied with respect to the DRP parameters:
\begin{align}
    \state{t} 
    &= \phi(\state{t-1}, \pi_\theta(\state{t-1}), \noise{t-1})\nonumber \\
    &= \phi(\phi(\state{t-2}, \pi_\theta(\state{t-2}), \noise{t-1}), \pi_\theta\left(\phi(\state{t-2}, \pi_\theta(\state{t-2}), \noise{t-1})\right), \noise{t-1}) \nonumber \\
    \label{eqn:full_reparam_to_h_drp}
    &= \dots \defeq \Psi_t(\state{h}, \theta, \noise{h:t-1}), \quad \forall t = h, h + 1, \dots H.
\end{align}
Now, the DRP can be readily computed by backpropagation through the entropic utility and (\ref{eqn:full_reparam_to_h_drp}):
\begin{align*}
 \pderiv{\hat{U}_h(\action{h:H})}{\theta} 
    &\propto \pderiv{}{\theta}\E{\state{h+1:H}}{\exp{\left(\beta \sum_{t=h}^H \reward{\state{t}}{\pi_\theta(\state{t})}\right)}} \\
    &= \E{\noise{h:H}}{\pderiv{}{\theta}\exp{\left(\beta \sum_{t=h}^H \reward{\Psi_t(\state{h}, \theta, \noise{h:t-1})}{\pi_\theta\left(\Psi_t(\state{h}, \theta, \noise{h:t-1}) \right)}\right)}}.
\end{align*}
In fact, (\ref{eqn:utility_value}) always attains a maximum for \emph{some} reactive policy \citep{bauerle2014more}, making the corresponding optimization problem well-defined. Also, the optimal policy $\pi_t^*$ can be well-approximated provided that $\pi_\theta$ is a sufficiently expressive class of function approximators.

\subsection{RAPTOR}
\label{ssec:raptor}

The model defined by these computations can be formally described by a \emph{stochastic computation graph} \citep{bueno2019deep,schulman-stochastic-gradient}. 
Here, the set of inputs to the graph includes the initial state $\state{0}$, actions $\action{0}, \dots \action{H}$, and $\beta$. The transitions $\state{t+1}=\phi(\state{t}, \action{t}, \noise{t})$ can be encoded using three edges: $\state{t}\rightarrow \state{t+1}$, $\action{t}\rightarrow \state{t+1}$, and $\noise{t} \rightarrow \state{t+1}$, in which $\noise{t}$ are stochastic nodes in the graph. Similarly, the rewards $r_t = \reward{\state{t}}{\action{t}}$ can be represented as leaf nodes in the graph, and described by using two dependencies: $\state{t}\rightarrow r_t$ and $\action{t}\rightarrow r_t$. From each reward node $r_t$, an edge extends to the sufficient statistics $\hat{\mu}$ and $\hat{\sigma}^2$ approximating utility $\utility{\noise{0:H}}{\sum_{t=0}^{H} r_t}$, in which the reparameterization trick has effectively transformed the randomness in $\state{h:H}$ to purely exogeneous noise $\noise{h:H}$. 

\begin{figure}[t!]
    \centering
     \begin{subfigure}[c]{0.5\textwidth}
        \centering
            \includegraphics[width=0.99\linewidth]{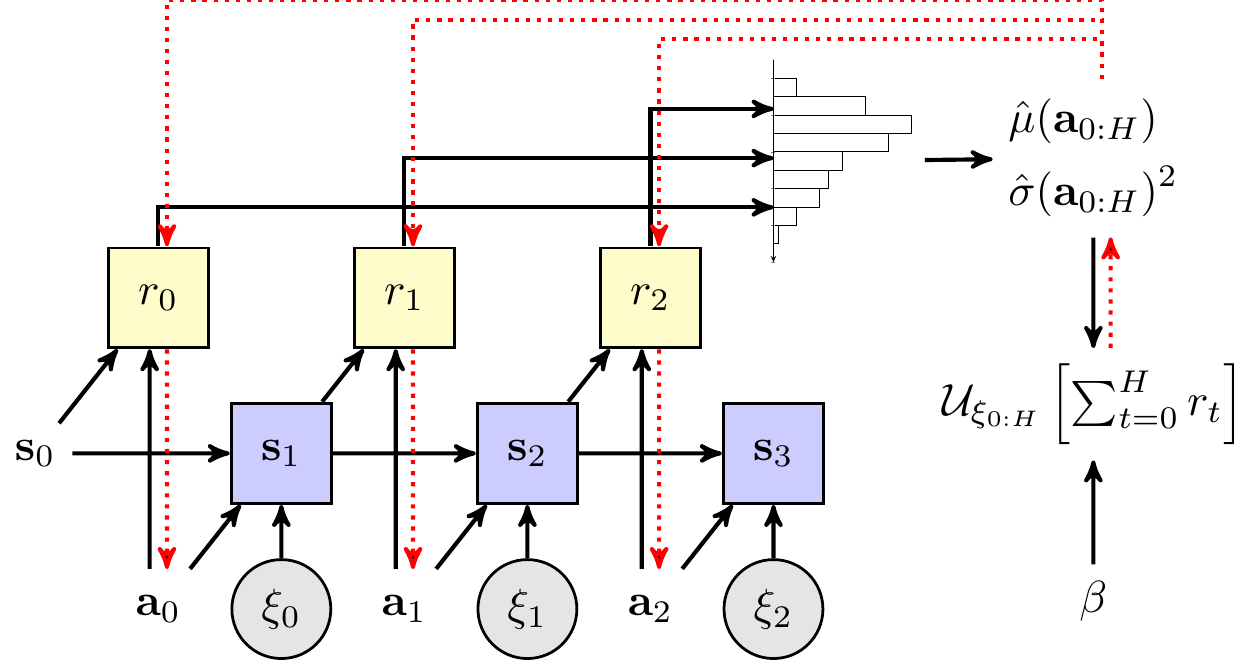}
        \caption{\acronym-SLP}
    \end{subfigure}%
    \begin{subfigure}[c]{0.5\textwidth}
        \centering
            \includegraphics[width=0.99\linewidth]{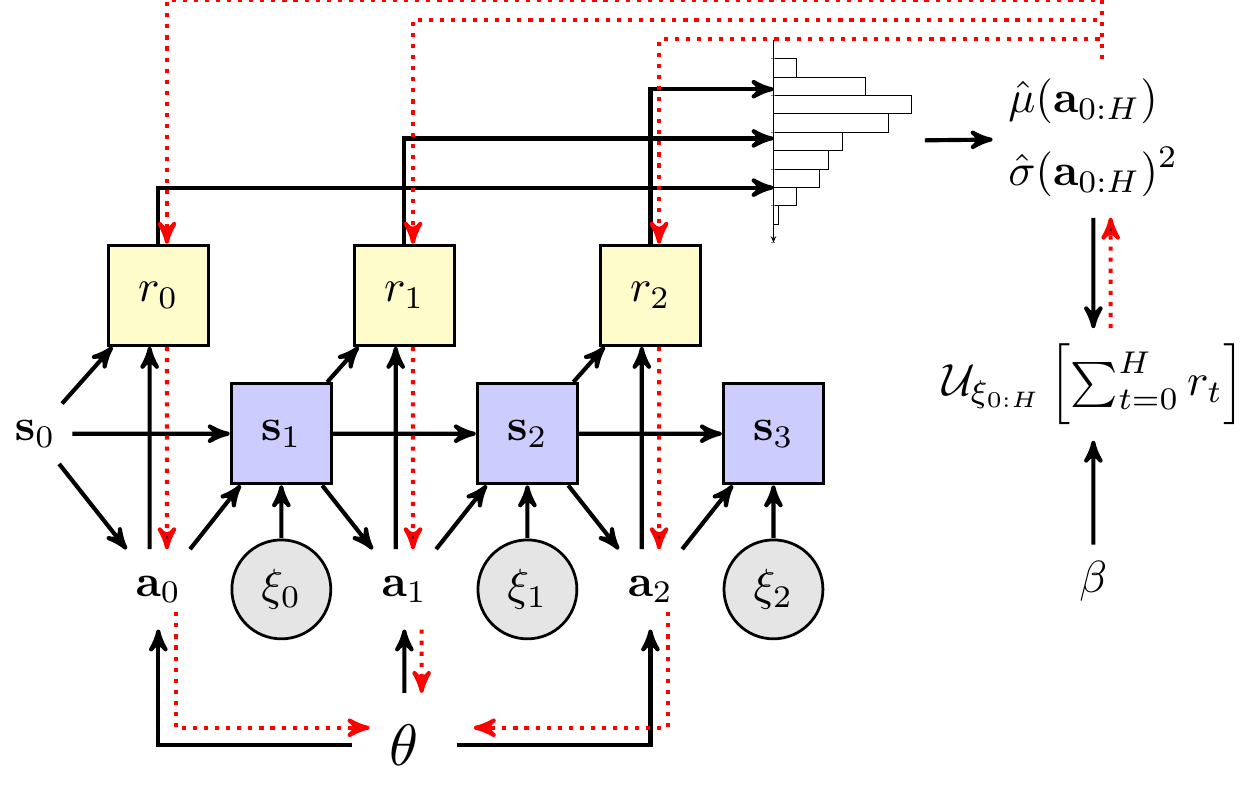}
        \caption{\acronym-DRP}
    \end{subfigure}%
    \caption{
    The stochastic computation graph of \acronym~for three decision steps. Following \citet{schulman-stochastic-gradient}, square and rounded nodes show deterministic and stochastic nodes, respectively. The input nodes are drawn without borders. Note that all state nodes $\state{1}, \state{2}\dots$ are deterministic due to the reparameterization via $\noise{t}$. During the forward pass, the inputs of the model along with a batch of samples of $\noise{t} \sim \noisedensity$ induce an empirical distribution (histogram) over $\sum_{t=0}^{H}\reward{\state{t}}{\action{t}}$. From this, we approximate the utility objective $\utilityfunction$, which is a symbolic function of the samples and their sufficient statistics. We show the flow of gradients during backpropagation in red dotted lines.}
    \label{fig:stochastic-computation-graph}
\end{figure}

Based on the results in Section \ref{ssec:stochastic-graph} and Section \ref{ssec:drp}, we can now define two possible instantiations of \acronym. The first approach computes an optimal straight-line plan by backpropagating through sample approximations of the mean and variance of returns, which we refer to as \acronym-SLP. The second approach approximates an optimal (deep) reactive policy, and we refer to this as \acronym-DRP. The computation graph defining both instances of \acronym~is depicted in Figure~\ref{fig:stochastic-computation-graph}, where edge dependencies are drawn as solid black arrows and gradients are drawn as dashed red arrows.

\section{Experiments}
\label{sec:experiments}
This section describes three domains which are used to analyze the performance of \acronym-SLP and \acronym-DRP. Additionally, the results and performance of \acronym~are compared to risk neutral agents for each of the three domains.

\subsection{Domain Descriptions}
\paragraph{Navigation} 
Navigation \citep{navigation-faulwasser} is a two-dimensional path-finding domain in which the goal is to find a shortest path from a fixed starting point to a fixed goal region. The continuous state $\state{t} = (s_{t,x}, s_{t,y})$ represents the location of the agent at time $t$ and the bounded continuous action $\action{t} = (a_{t,x}, a_{t,y}) \in [-b, b]^2$ is the coordinate displacement of the agent at time $t$ for some $b>0$. The agent's objective is to reach the goal region in a minimal number of time steps. Thus, the reward at each time step can be defined as the negative Euclidean distance from the agent to the center of the goal region $\mathbf{g} = (g_x, g_y)$, e.g. $\reward{\state{t}}{\action{t}} = -\| \state{t} - \mathbf{g}\|_2$. Notably, next state transitions are subject to high variability when the agent passes through a high-variance zone in between the starting point and the goal region. In particular, let $\mathrm{crossing}_t$ denote the length of a sub-trajectory that crosses the zone when moving from $\state{t}$ to $\state{t}+\action{t}$. Then, a normally distributed noise $\mathcal{N}(0, (\mathrm{crossing}_t\sigma_h)^2)$ is added to the next state. On the other hand, when $\mathrm{crossing}_t=0$, a small noise $\mathcal{N}(0,\sigma_l^2)$, whose variance $\sigma_l^2 \ll \sigma_h^2$, is added.
Thus, the transition function for the navigation domain can be reparameterized as:
\begin{equation*}
    \phi(\state{t}, \action{t}, \noise{t}) = 
        \state{t} + \action{t} + \mathrm{crossing}_t \sigma_h\noise{t} + \mathbbm{1}[\mathrm{crossing}_t = 0] \sigma_l\noise{t}, \quad \noise{t}\sim \mathcal{N}(\mathbf{0}, \mathbf{I}_{2\times 2}).
\end{equation*}

\paragraph{Reservoir Control}
The reservoir control domain \citep{yeh1985reservoir} involves controlling the flow of water between $N = 5$ interconnected reservoirs. The water level of each reservoir $i$ at time $t$ is represented by $s_{t,i} \in \reals_+$. The action $a_{t,ij}=[\action{t}]_{ij}$ is the amount of water discharged from reservoir $i$ to $j$ for $j\in \mathrm{downstream}(i)$, where $\mathrm{downstream}(i)$ is the set of downstream reservoirs of $i$ (we define $\mathrm{upstream}(i)$ similarly). 
The action is constrained such that $a_{t, ik}=0,\forall{k}\notin \mathrm{downstream}(i)$. Also, 
the total outflow of a reservoir cannot exceed the reservoir's current water level, i.e., $\sum_j a_{t, ij} \leq s_{t,i}$. The goal of the agent is to keep the water level within a safe range $[L_i, U_i]$, so we define the reward function as:
\begin{equation*}
    \begin{aligned}
        R_i \defeq 
        \begin{cases}
            -P_u (s_{t,i} - U_i),    &\text{if } s_{t,i} \geq U_i \\
            -P_l (L_i - s_{t,i}), &\text{if } s_{t,i} \leq L_i \\
            0,                      &\text{otherwise}  \\
        \end{cases}, \qquad 
        \reward{\state{t}}{\action{t}} = \sum_{i = 1}^N R_i,
    \end{aligned}
\end{equation*}
where $P_u$ ($P_l$) is the penalty for water levels above (below) the upper (lower) bound. Note that $P_u$ is weighted more heavily than $P_l$, since overflows can lead to costly flooding and damages
, whereas shortages can more easily be resolved by, e.g., supplementing water from secondary sources. The amount of rainfall at each time step is modelled as an exponentially-distributed random variable $\noise{t}$ with rate $\lambda$. Thus, switching to a component-wise view of $\phi_i$ for notational clarity, 
the transition function for each reservoir $i$ becomes

\begin{equation*}
    \phi_i(\state{t}, \action{t}, \noise{t}) = s_{t,i} -\sum_{j \in \mathrm{downstream}(i)}a_{t,ij} + \sum_{j \in \mathrm{upstream}(i)}a_{t,ji} + \noise{t}.
\end{equation*}

\paragraph{HVAC Control}
In the HVAC control domain, an agent modulates the volume of heated air flow into each of the $N = 5$ rooms in a house. The state consists of the temperature in each room $i$ given by $s_{t,i} \in \reals$. The air volume, $\action{t}$, into each room is non negative ($a_{t,i} \geq 0$) with constant temperature $T_c$. The objective is to keep the rooms at their set temperature and to avoid them getting too cold, so the reward function can be defined as:
\begin{equation*}
    \begin{aligned}
        R_i \defeq
        \begin{cases}
            -|s_{t,i} - T_i| - a_{t,i} - P_u,    &\text{if } s_{t,i} \leq T_l \\
            -|s_{t,i} - T_i| - a_{t,i}, &\text{otherwise} \\
        \end{cases}, \qquad
        \reward{\state{t}}{\action{t}} = \sum_{i=1}^N R_i,
    \end{aligned}
\end{equation*}
where $T_i$ is the set temperature for room $i$, $T_l$ is a threshold on the temperature below which it would be considered too low, and $P_u$ is the penalty for the temperature going below $T_l$. As a cost saving measure, $T_i$ is set near $T_l$ to avoid expensive heating. Stochasticity is added in three ways for HVAC: (1) outdoor temperature is distributed as $\mathcal{N}(T_o, \sigma_o^2)$; (2) noise $\mathcal{N}(0, \sigma_a^2)$ in the volume of air applied to each room; and (3) a small variance $\mathcal{N}(0, \sigma_{ij}^2)$ in the heat dispersion between rooms. Combining this, the transition function for each room $i$ is:
\begin{equation*}
    \phi_i(\state{t},\action{t},\noise{t}) = s_{t,i} + a_{t,i}(T_c-s_{t,i}) + \sigma_a\noise{t} + \sum_{j\in \mathrm{Adjacent}(i)}\frac{(s_{t,j} - s_{t,i})^3}{\rho_{ij}} + \sigma_{ij}\noise{t} + T_o + \sigma_o\noise{t},
\end{equation*}
where $\noise{t}\sim \mathcal{N}(0,1)$, $\rho_{ij}$ is the thermal resistance between adjacent rooms $i$ and $j$ and $\mathrm{Adjacent}(i)$ are the rooms adjacent to $i$.

\subsection{Performance of Risk-Aware Planning}
\label{ssec:discussion}
\begin{figure}[h!]
    \centering
     \begin{subfigure}[c]{0.34\textwidth}
        \centering
            \includegraphics[width=0.99\linewidth]{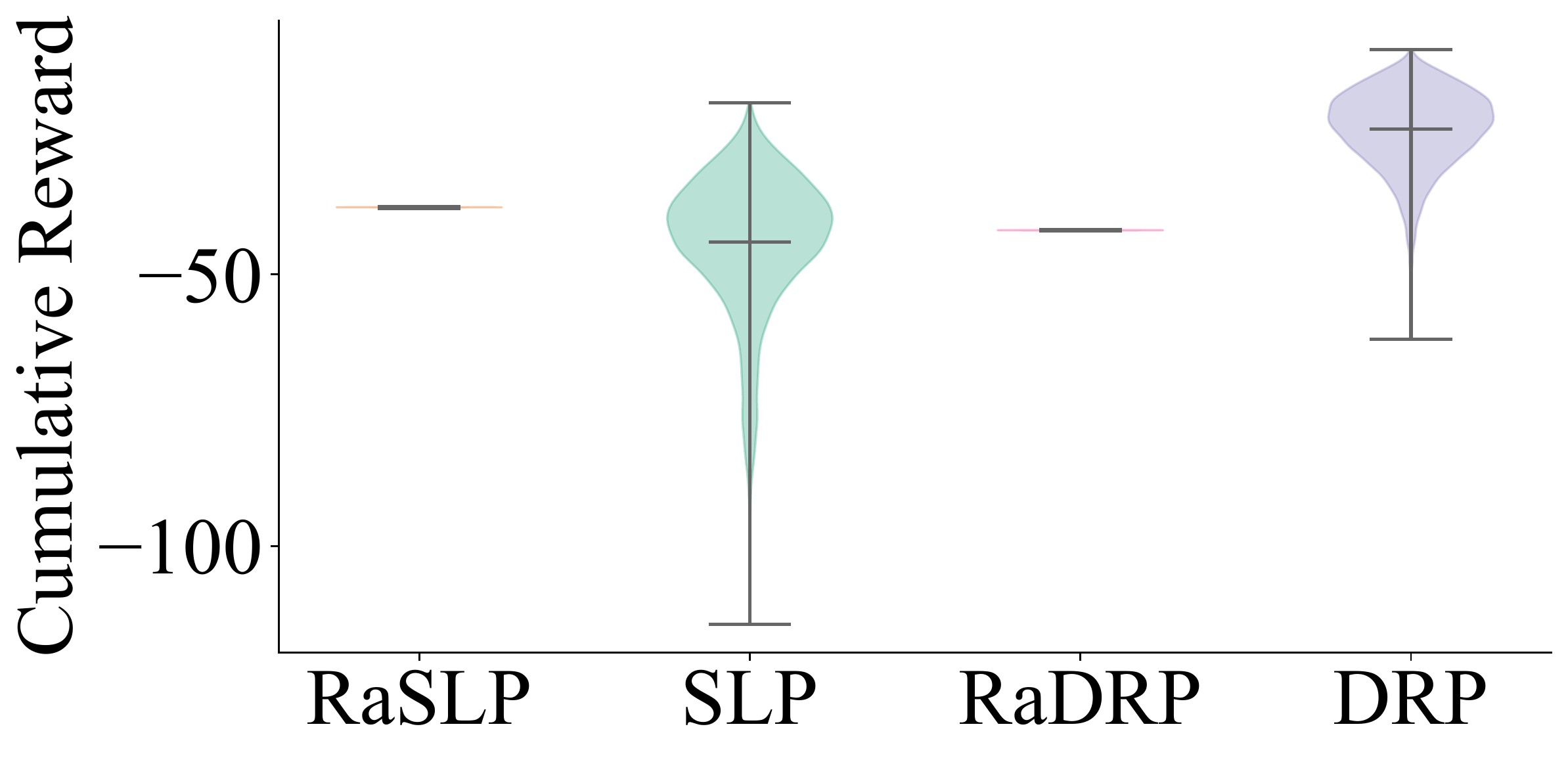}
        \caption{Navigation}
        \label{fig:cum-reward-nav}
    \end{subfigure}%
    \begin{subfigure}[c]{0.31\textwidth}
        \centering
            \includegraphics[width=0.99\linewidth]{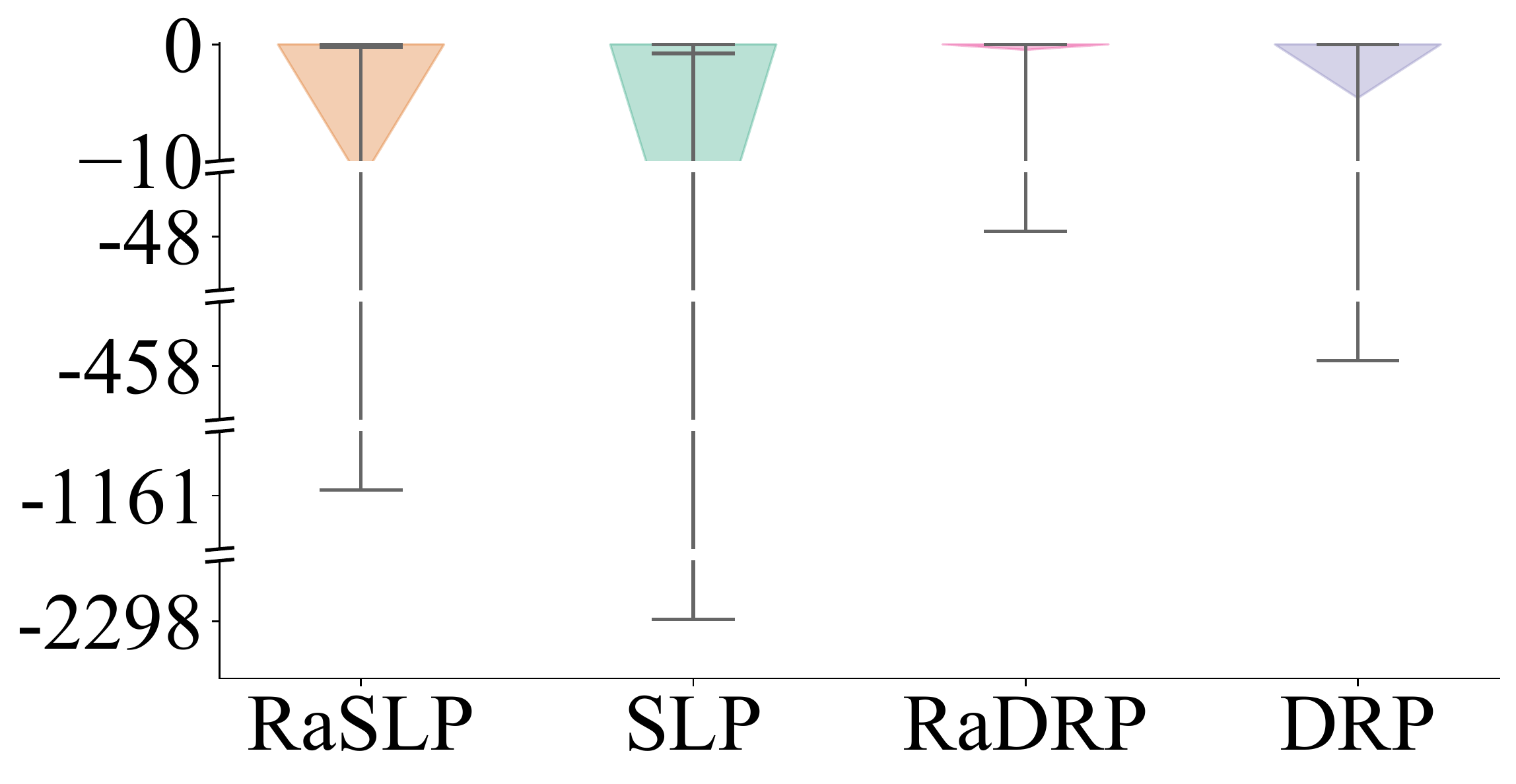}
        \caption{Reservoir}
        \label{fig:cum-reward-reservoir}
    \end{subfigure}%
    \begin{subfigure}[c]{0.35\textwidth}
        \centering
            \includegraphics[width=0.99\linewidth]{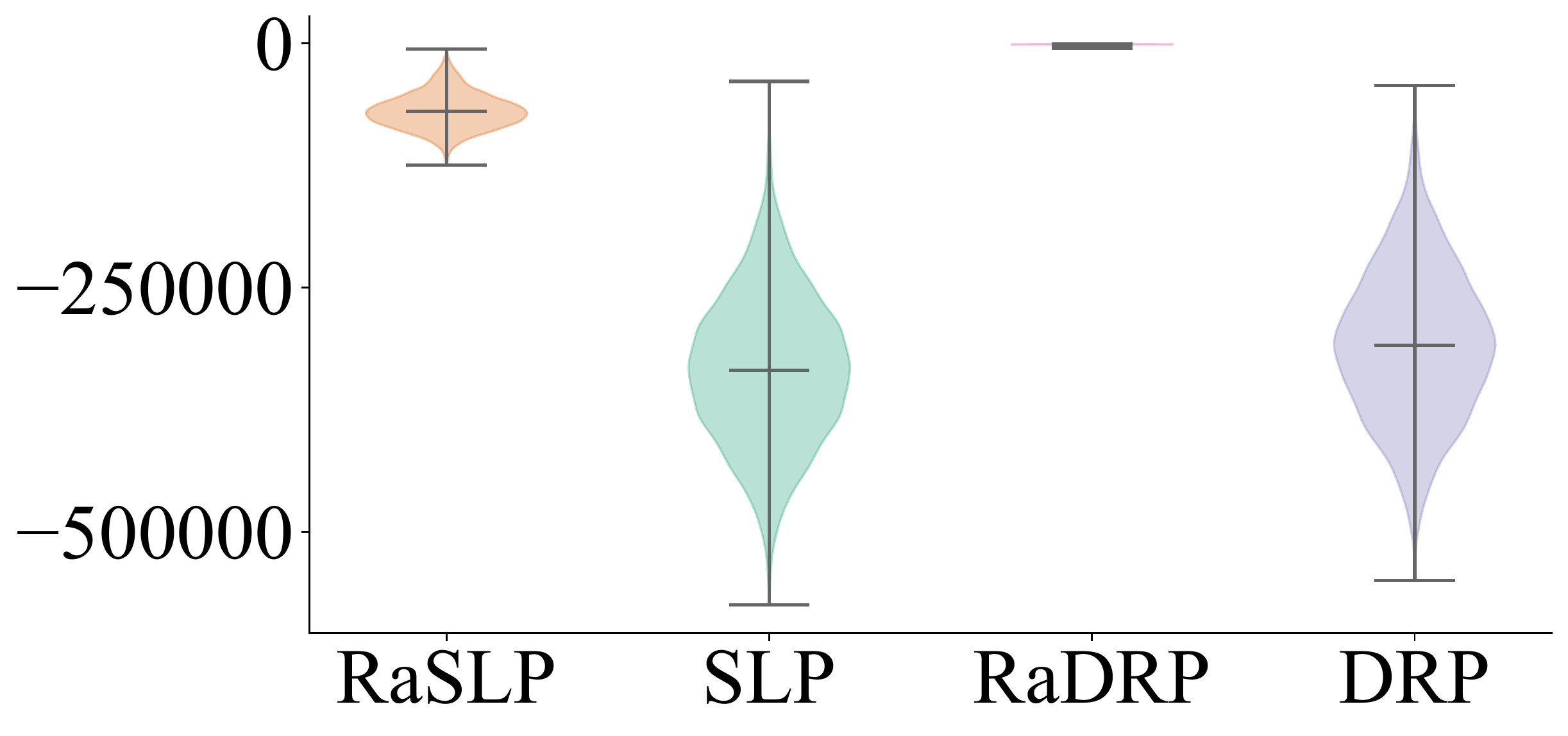}
        \caption{HVAC}
        \label{fig:cum-reward-hvac}
    \end{subfigure}%
    \caption{The cumulative reward distributions for all three domains are shown with SLP, \acronym-SLP, DRP and \acronym-DRP agents shown. To generate these distributions $10, \!000$, $250, \!000$ and $10, \!000$ trajectories were gathered for each agent in navigation, reservoir and hvac respectively.}
    \label{fig:cumulative_reward_dists}
\end{figure}

In this section, we analyze and compare the empirical performance of \acronym-SLP (RaSLP) and \acronym-DRP (RaDRP) against two recent scalable risk-neutral planners as baselines, the SLP approach in \citet{wu2017scalable} and the DRP in \citet{bueno2019deep}. Please note that relevant hyper-parameters used in the experiments were tuned specifically for the baselines, with the exception of $\beta$, and used by \acronym~for fair comparison (for detailed breakdown see Appendix \appendixref{ssec:detailed_experiment}{A.3}). The distributions of returns from all 3 domains are illustrated in Figure \ref{fig:cumulative_reward_dists}. Further analysis of the performance for different values of $\beta$ is provided in Appendix \appendixref{sec:ablation}{A.2}.

\begin{figure}[!h]
    \centering
    \includegraphics[width=1.0\linewidth]{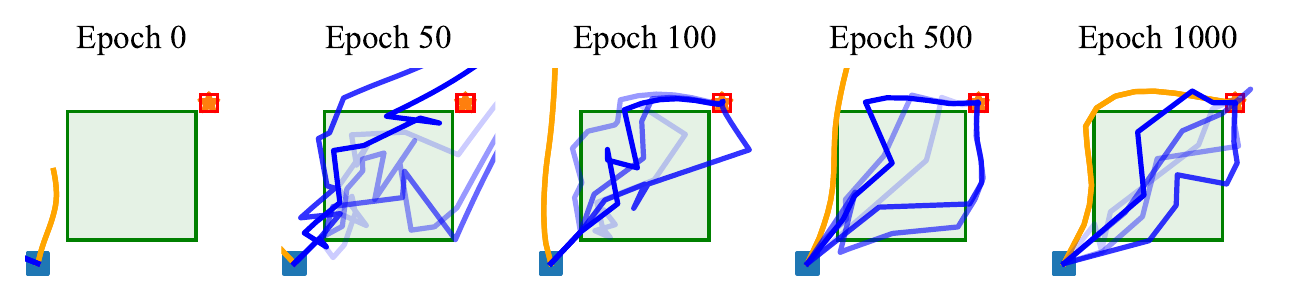}
    \caption{Similar to Figure~\ref{fig:new_nav_sidebyside}, the evolution of trajectories of \acronym-DRP (orange) and a risk-neutral-DRP agent (blue) are shown.}
    \label{fig:new_nav_sidebyside_drp}
\end{figure}

\paragraph{Navigation Results}
%
Figure~\ref{fig:cum-reward-nav} compares two \acronym\ agents against risk-neutral baselines. SLP achieves the lowest expected return as it cannot adapt to the highly stochastic environment and hence frequently fails to reach the goal region. DRP, on the other hand, attains the greatest expected return as it reactively approaches the goal region while passing through the high-variance zone (Figure~\ref{fig:new_nav_sidebyside_drp}). However, we can clearly see the advantage of risk-sensitive plans as they are able to obtain reasonably high returns while significantly reducing the variance in the returns at the same time. Figure~\ref{fig:new_nav_sidebyside} and Figure~\ref{fig:new_nav_sidebyside_drp} show that \acronym\ agents circumvent the high-variance zone to reliably achieve high returns. Overall, the \acronym\ agents are unequivocally preferred in this domain when we have to avoid the worst-case outcomes.
%

\begin{figure}[!h]
    \centering
     \begin{subfigure}[c]{0.2\textwidth}
        \centering
            \includegraphics[width=0.99\linewidth]{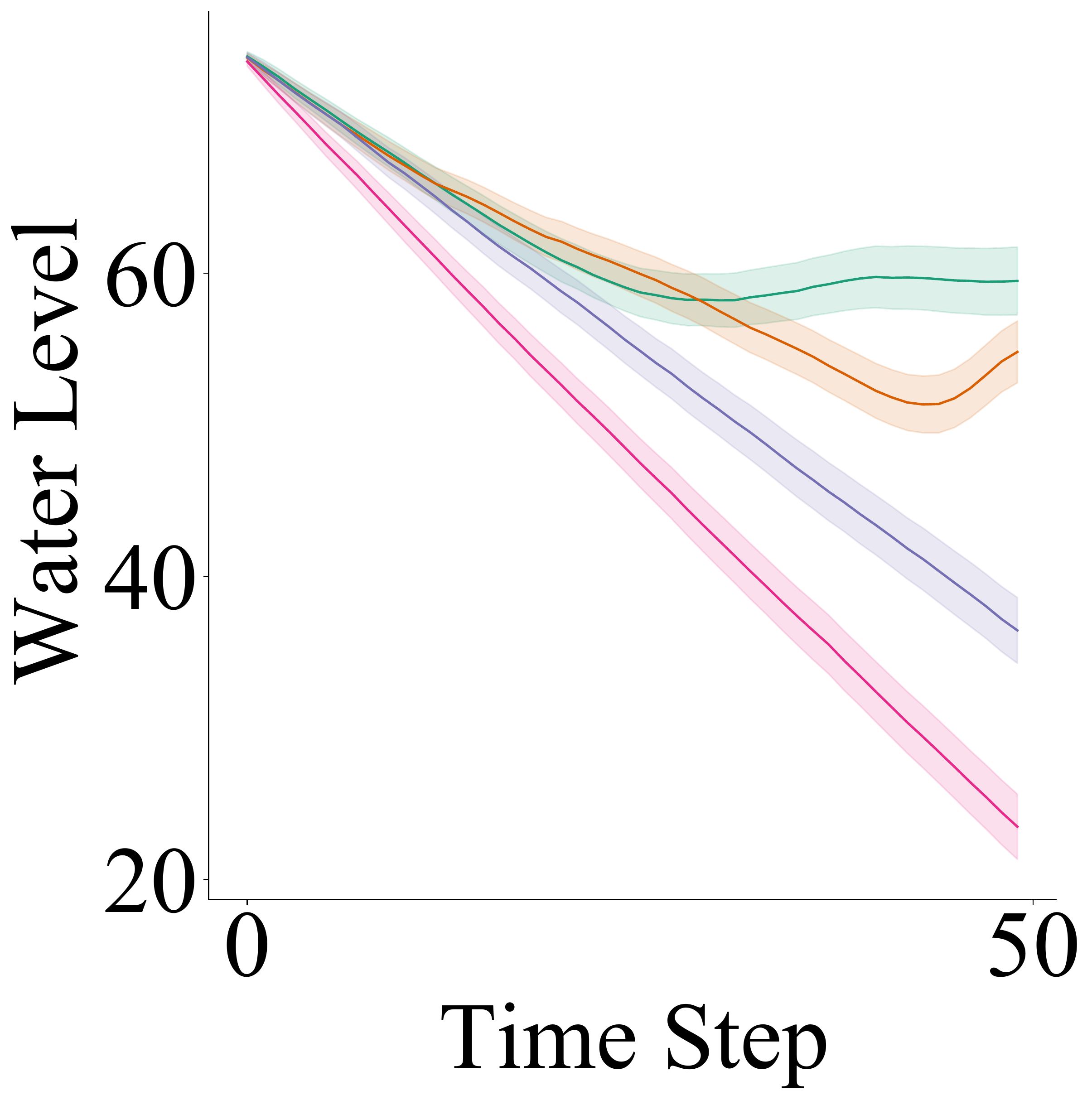}
        \caption{Reservoir 1}
    \end{subfigure}%
    \begin{subfigure}[c]{0.2\textwidth}
        \centering
            \includegraphics[width=0.99\linewidth]{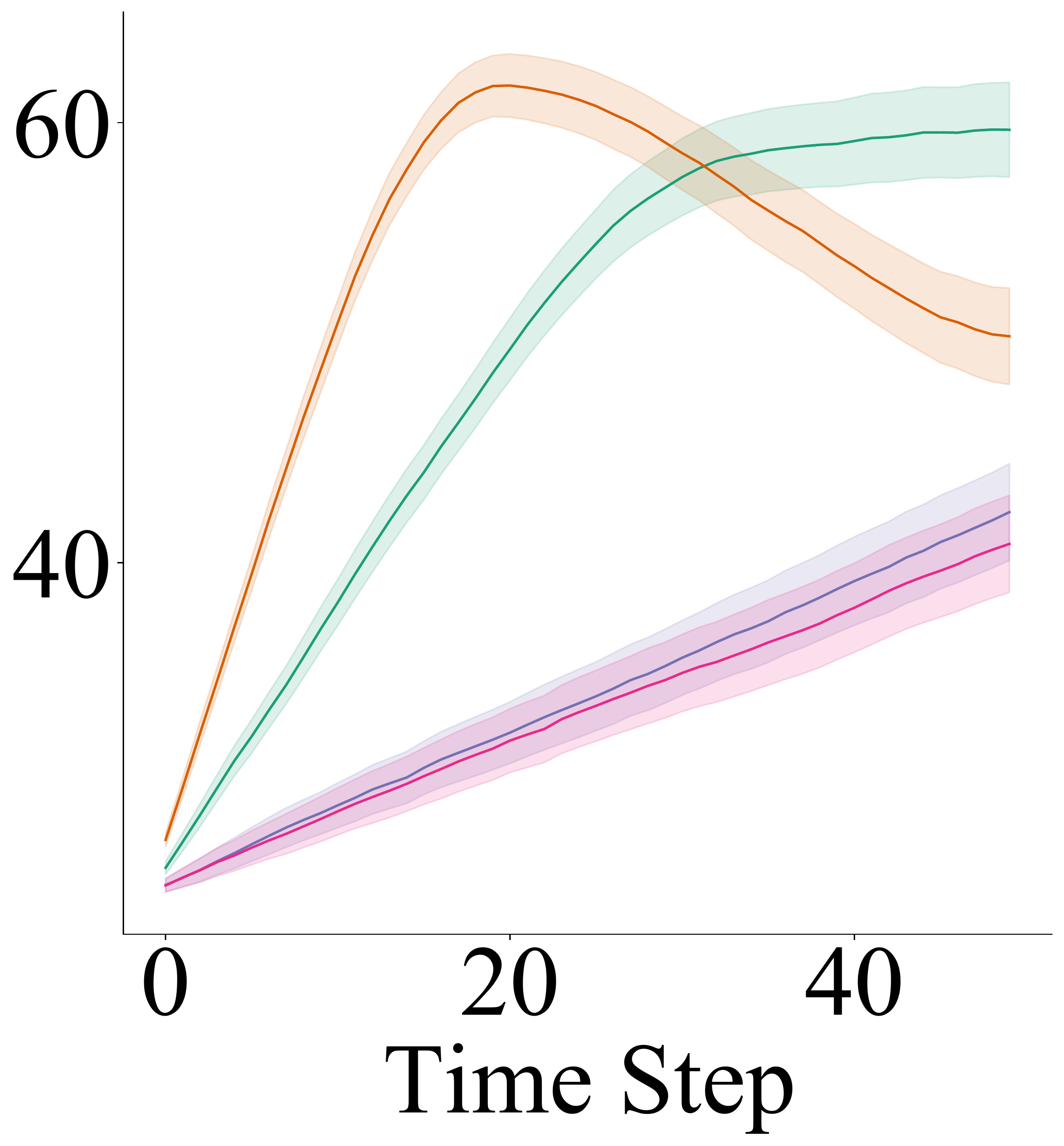}
        \caption{Reservoir 2}
    \end{subfigure}%
    \begin{subfigure}[c]{0.2\textwidth}
        \centering
            \includegraphics[width=0.99\linewidth]{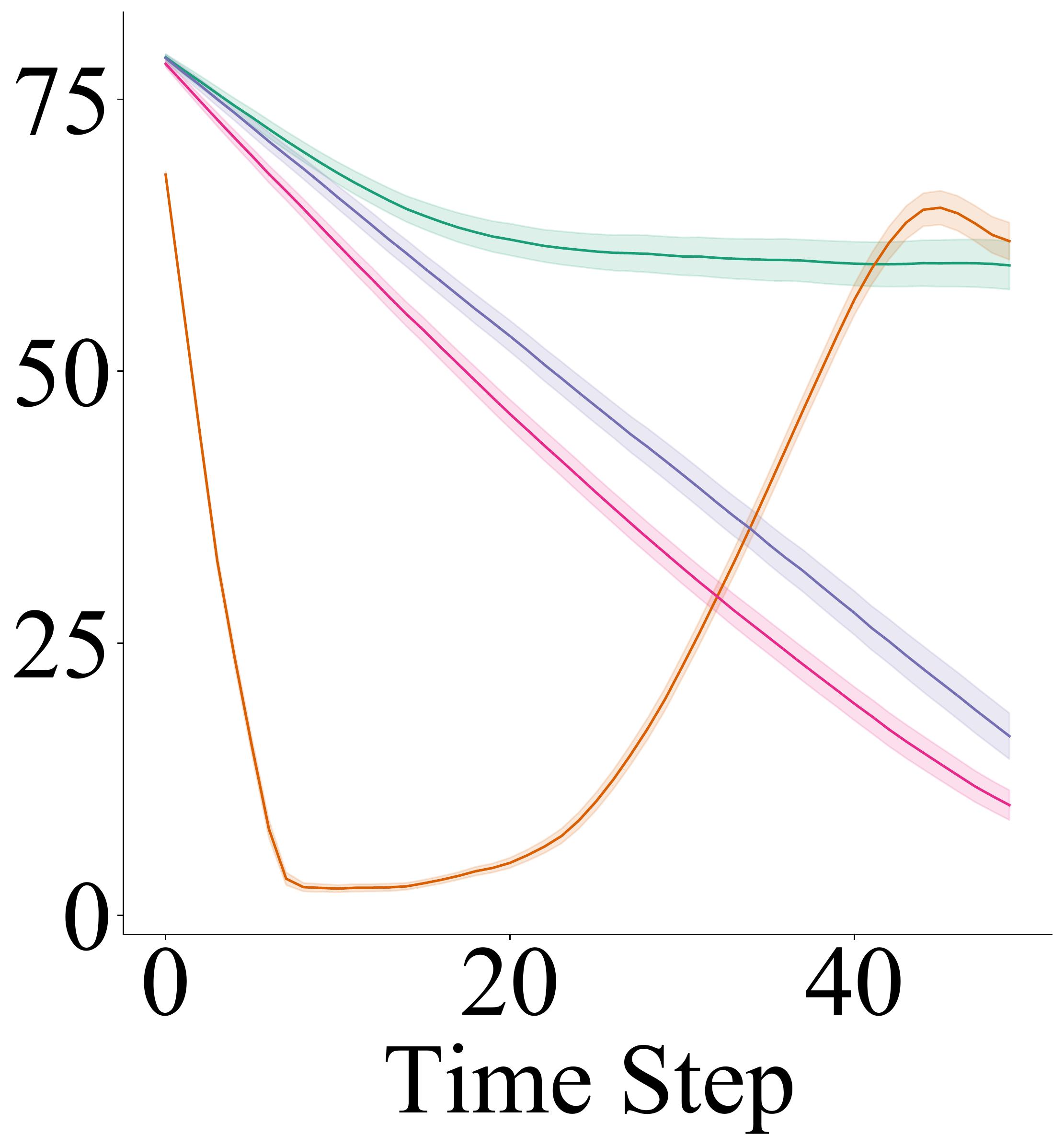}
        \caption{Reservoir 3}
    \end{subfigure}%
    \begin{subfigure}[c]{0.2\textwidth}
        \centering
            \includegraphics[width=0.99\linewidth]{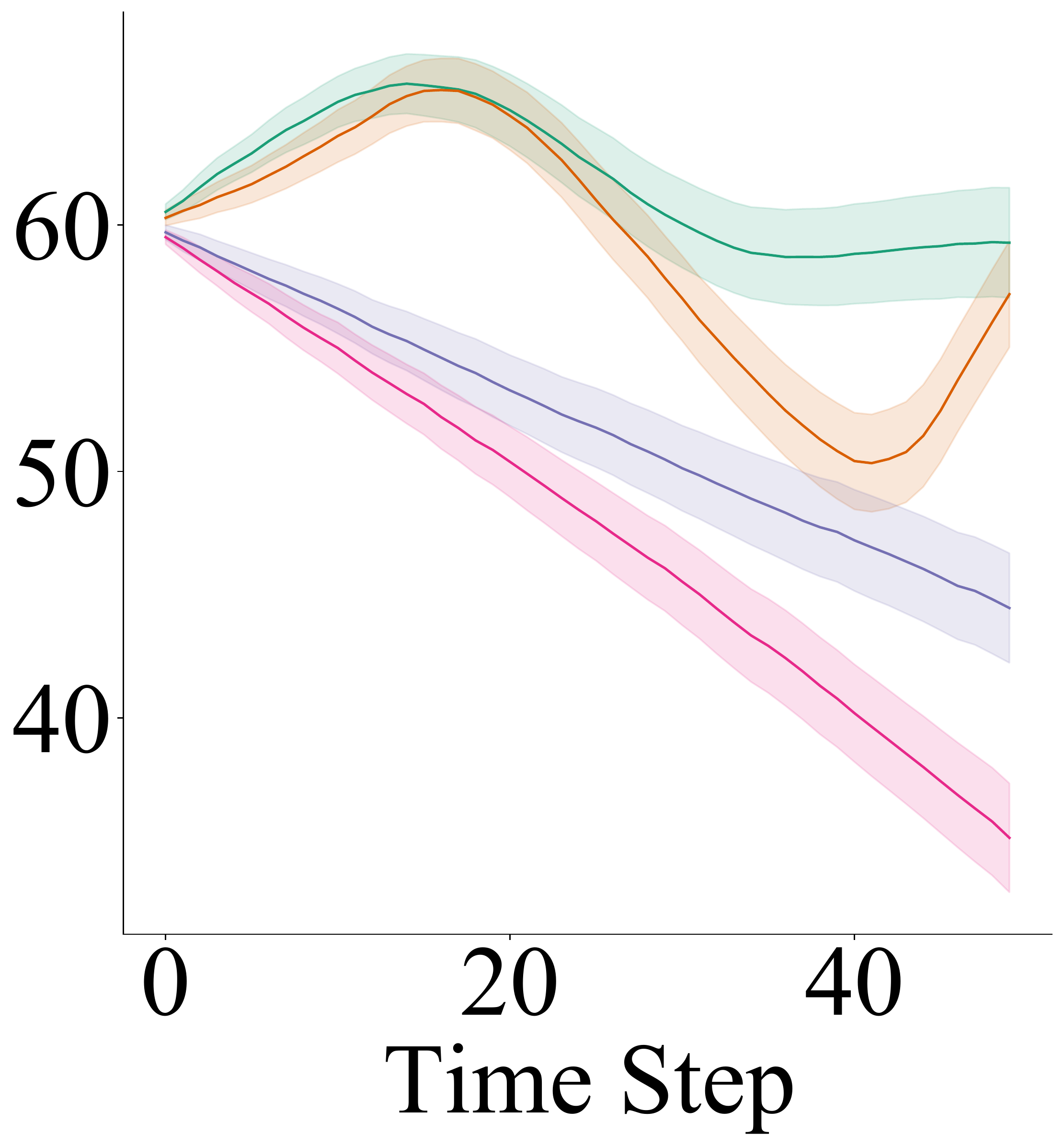}
        \caption{Reservoir 4}
    \end{subfigure}%
    \begin{subfigure}[c]{0.2\textwidth}
        \centering
            \includegraphics[width=0.99\linewidth]{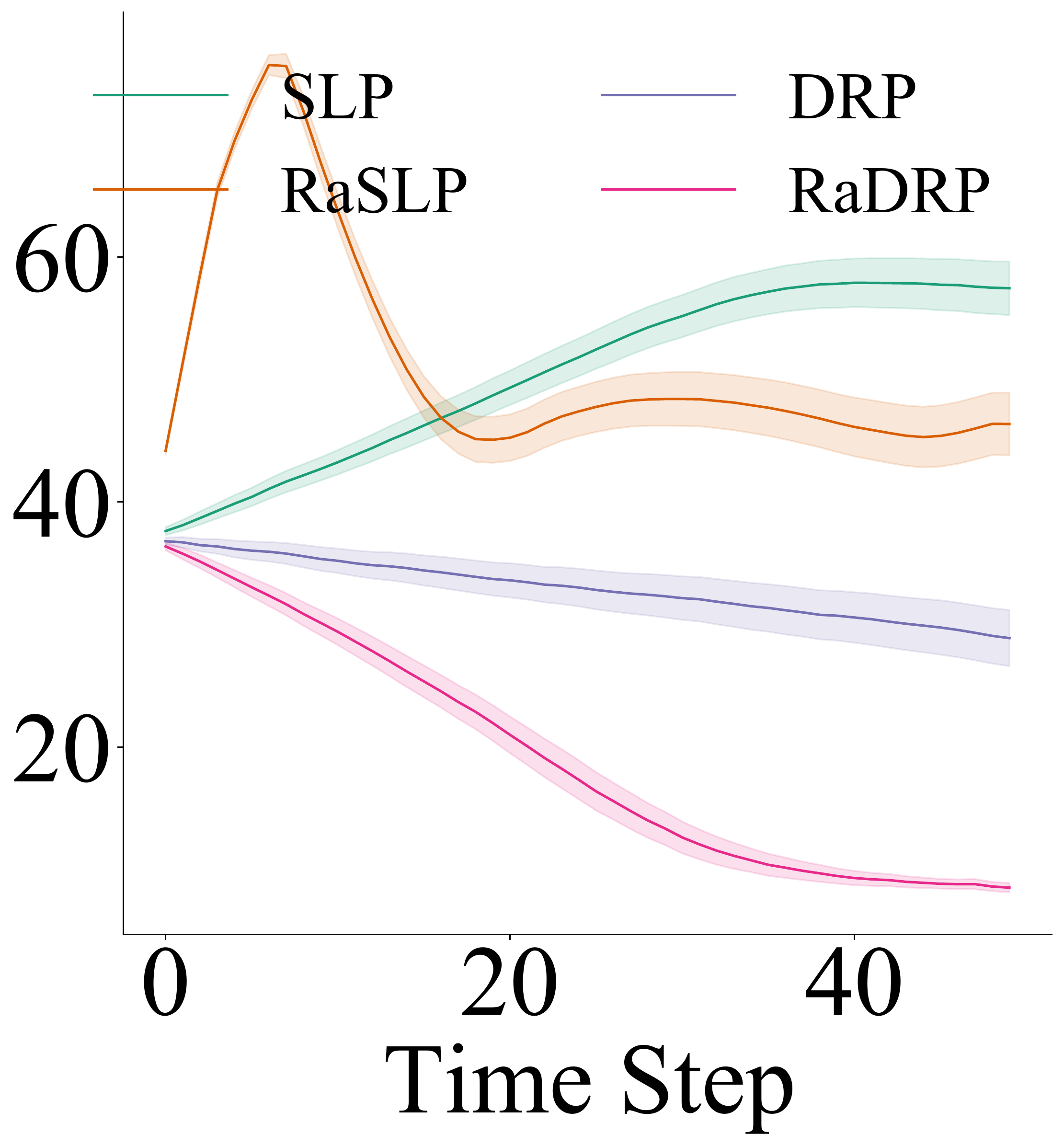}
        \caption{Reservoir 5}
    \end{subfigure}%
    \caption{The mean water levels and associated noise is shown for each of the five rooms with SLP, \acronym-SLP, DRP and \acronym-DRP.}
    \label{fig:reservoir_water_levels}
\end{figure}

\paragraph{Reservoir Results}
We can see in Figure~\ref{fig:cum-reward-reservoir} that both \acronym\ agents have smaller variances in the return than their risk-neutral counterparts. Figure~\ref{fig:reservoir_water_levels} shows that the optimized plan for the \acronym-SLP tends to maintain lower water levels than that for the SLP. Thus, \acronym\ agent is less likely to incur the large overflow penalty. Although DRP is able to attain smaller variance than \acronym-SLP by properly reacting to unveiled state transitions, further reduction in variance seen by \acronym-DRP is substantial. Also, we observe that the \acronym\ agents gain higher expected returns than the risk-neutral agents. This is because penalties are incurred only when the water levels go outside of the safe bounds; hence, keeping the levels within the bounds all the time ensures higher expected return as well. This shows the potential of \acronym\ in safety-critical applications.

\begin{figure}[!h]
    \centering
     \begin{subfigure}[c]{0.2\textwidth}
        \centering
           \includegraphics[width=0.99\linewidth]{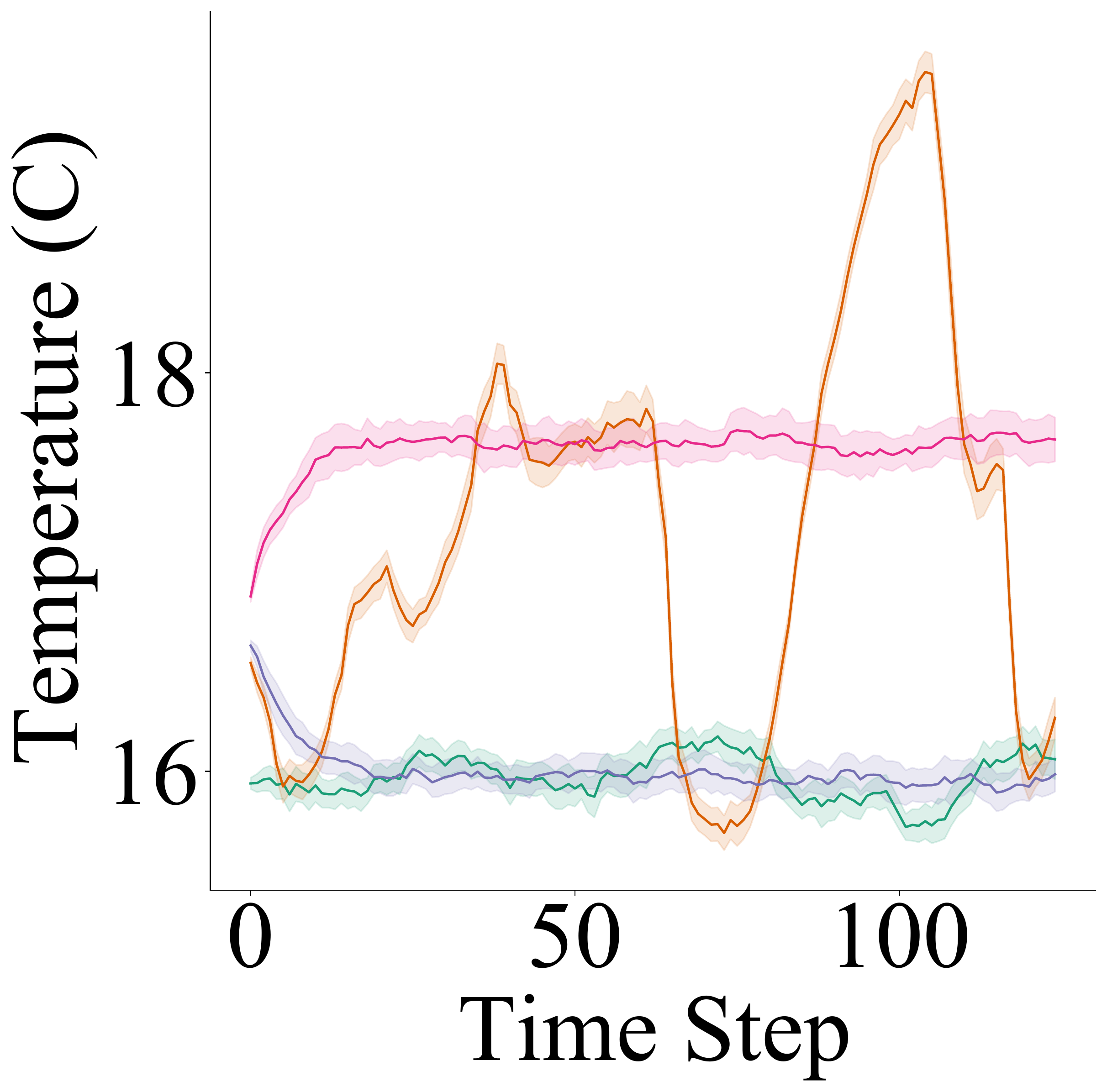}
        \caption{Room 1}
    \end{subfigure}%
    \begin{subfigure}[c]{0.2\textwidth}
        \centering
        \includegraphics[width=0.99\linewidth]{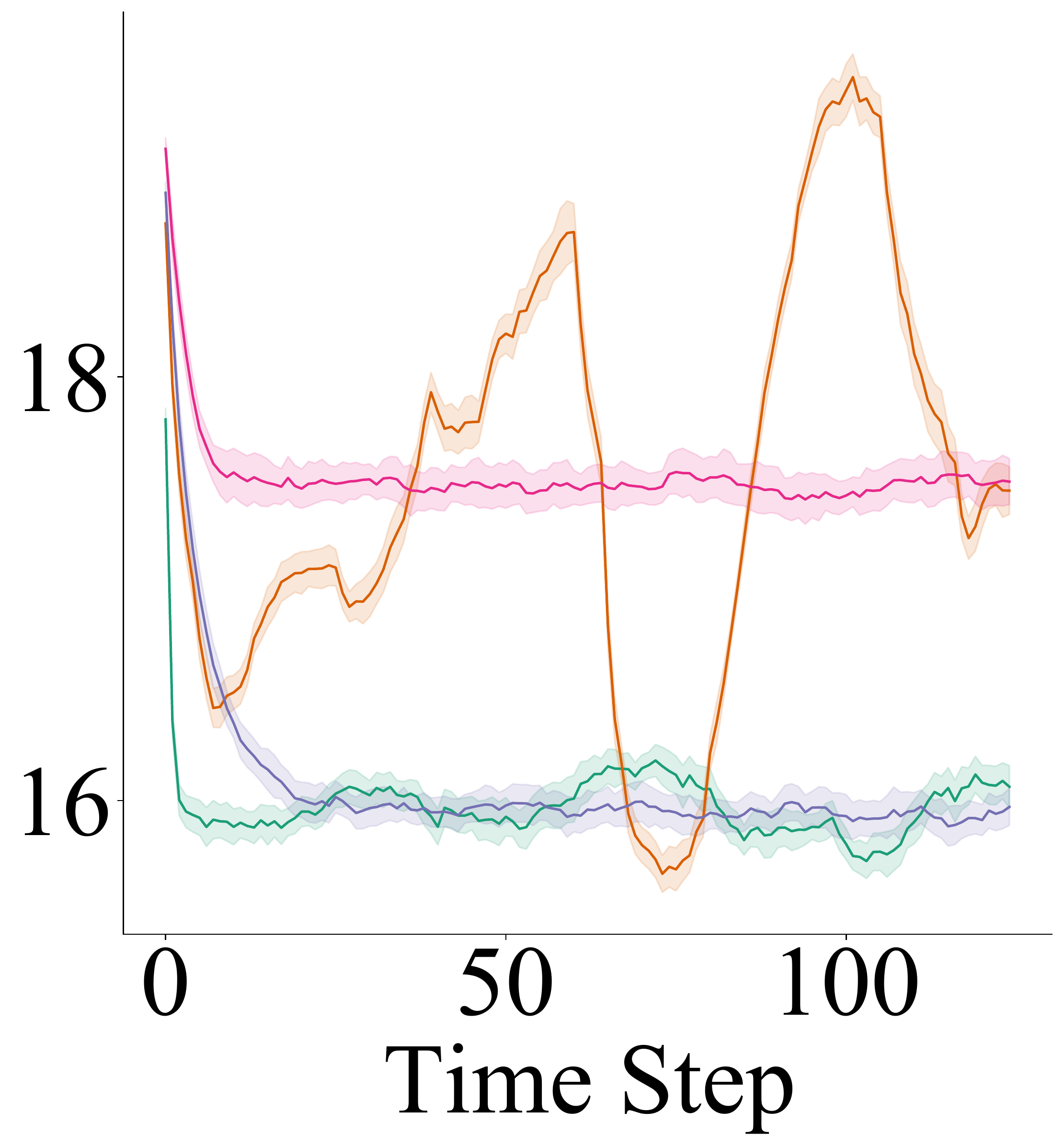}
        \caption{Room 2}
    \end{subfigure}%
    \begin{subfigure}[c]{0.2\textwidth}
        \centering
           \includegraphics[width=0.99\linewidth]{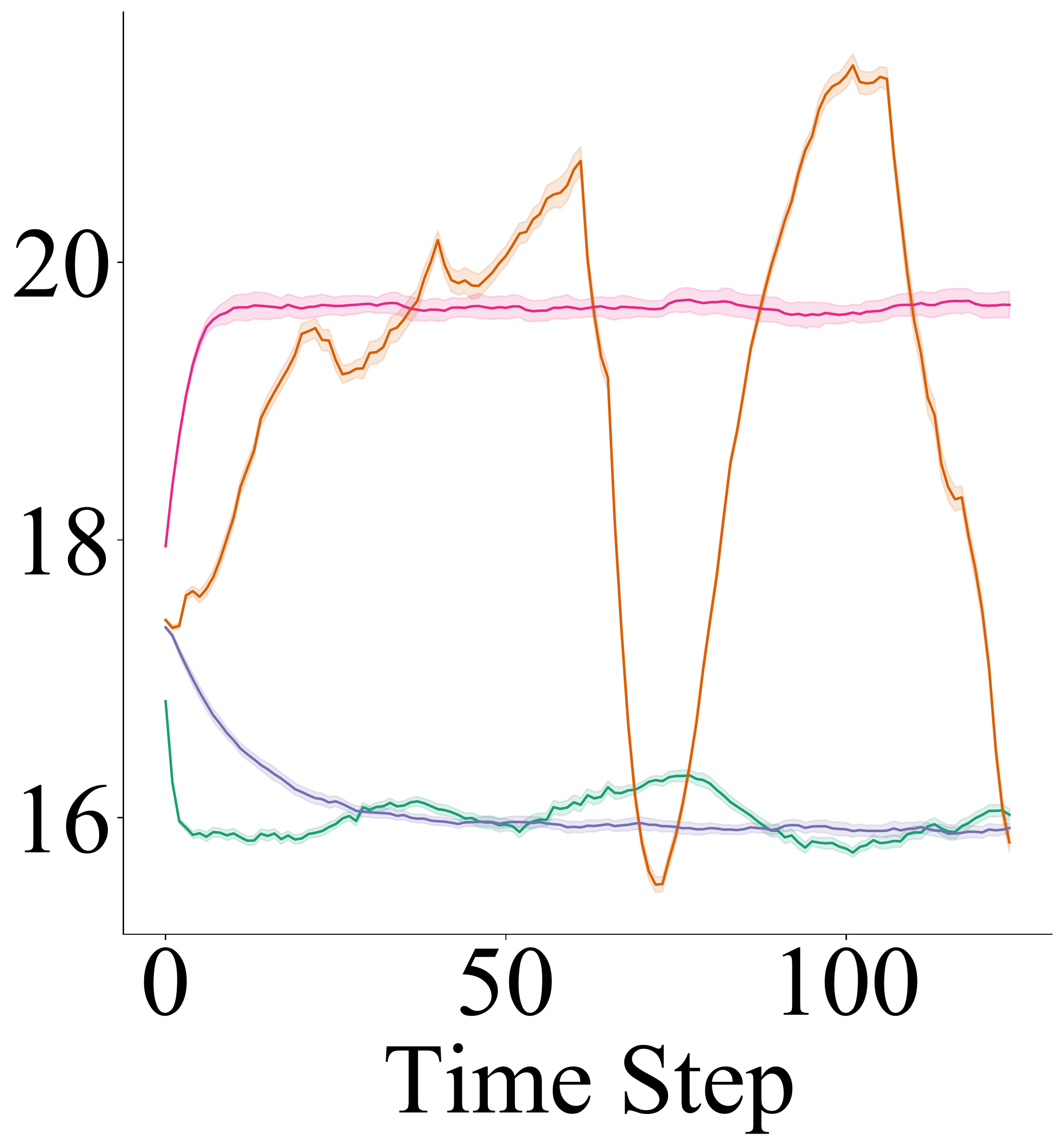}
        \caption{Room 3}
    \end{subfigure}%
    \begin{subfigure}[c]{0.2\textwidth}
        \centering
           \includegraphics[width=0.99\linewidth]{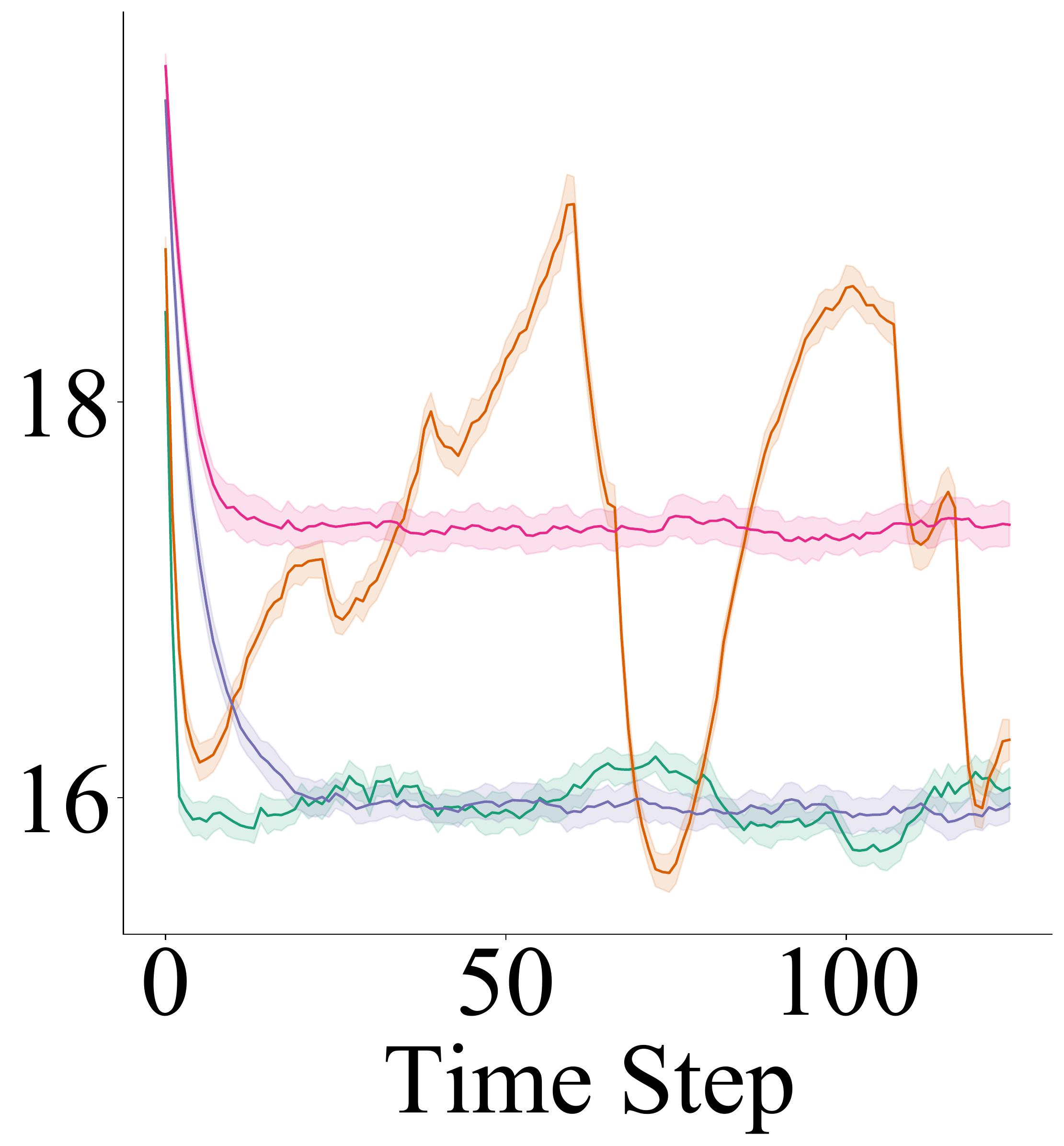}
        \caption{Room 4}
    \end{subfigure}%
    \begin{subfigure}[c]{0.2\textwidth}
        \centering
           \includegraphics[width=0.99\linewidth]{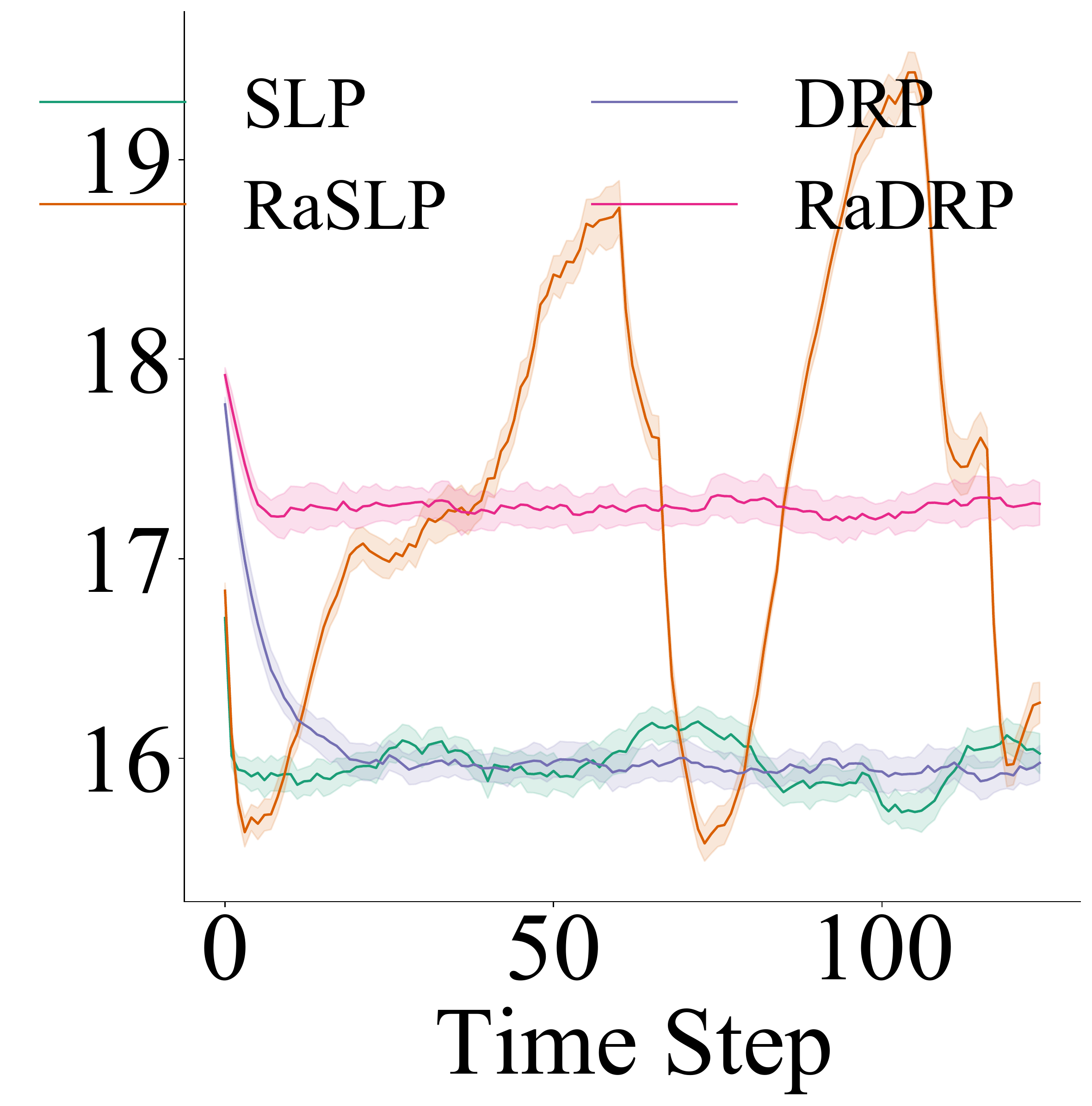}
        \caption{Room 5}
    \end{subfigure}
    \caption{The mean room temperatures and associated noise is shown for each of the five rooms with SLP, \acronym-SLP, DRP and \acronym-DRP.}
    \label{fig:hvac_room_temps}
\end{figure}

\paragraph{HVAC Results}
Figure~\ref{fig:cum-reward-hvac} shows that both \acronym\ agents obtain lower variance in the return than their respective risk-neutral agents. 
\acronym-DRP accomplishes this by setting the room temperatures consistently high so that the penalty $P_u$ can be avoided at all times (Figure~\ref{fig:hvac_room_temps}). Compared to \acronym-DRP, the optimized plan for \acronym-SLP induces more variability in the temperatures over time; however, the agent is still able to keep the temperatures from going below $T_l$ most of the times. In contrast, risk-neutral agents try to set the temperatures near $T_l$ to minimize the energy cost, which often leads to large penalties in this highly stochastic environment. Thus, the risk-neutral agents attain lower expected returns similar to the results in the Reservoir domain.

\section{Conclusion and Future Work}
\label{sec:conclusion}

We proposed \acronym, a scalable end-to-end risk-aware planner based on gradient descent applied to a risk-sensitive utility function. To this end, we extended \tfplan\ in order to accommodate stochastic transitions, by representing the planning problem as a stochastic computation graph and applying the reparameterization trick. Critically, we introduced the entropic utility as objective function, which can be seamlessly embedded in the graph as a symbolic objective. Then, \emph{risk-aware planning can be done end-to-end} using an off-the-shelf automatic differentiation tool (e.g., PyTorch) by exploiting samples of the utility and their sufficient statistics. We applied these techniques to learn a straight-line plan which commits to a sequence of actions in advance, as well as deep reactive policies that can better adapt to highly stochastic environments. Experiments on three highly stochastic domains --- Navigation, Reservoir, and HVAC -- demonstrated the ability of \acronym~to learn meaningful risk-sensitive policies in stochastic environments while maintaining scalability.

As future work, 
we note that extending our current approach to handle hybrid (mixed continuous and discrete) MDPs should be straightforward by using tricks such as projected gradients. Also, measuring and controlling risk is a complex topic in sequential decision making. The entropic utility objective and its mean-variance approximation may not capture all possible notions of risk. Applying this approach in situations such as financial portfolio optimization, where a model of dynamics is difficult to specify or even learn, and where the risk requirements could be potentially different, could lead to misrepresentation of risk. On the other hand, this could be addressed in different ways by using other well-known utility functions with different properties, such as CVaR \citep{chow2015risk}. However, this could be more difficult to incorporate into planning due to their reliance on percentiles of the return distribution and lack of asymptotic expansions exploited in this work. Incorporating other notions of risk could form an interesting and challenging future extension of our work. Another interesting extension could integrate model uncertainty for doing robust risk-aware planning under model misspecification.

\small
\bibliographystyle{plainnat}
\bibliography{references}
\normalsize

\ifappendix
\clearpage
\appendix

\section{Appendix}

\subsection{Proof of Theorem \ref{thm:slp}}
\label{subsec:proof}

Let $\xi$ be a random variable, $A$ be an arbitrary set and $X_a = g_a(\xi)$ for $a \in A$ and some functions $g_a$. Since $\prob{\sup_{a\in A} X_a \geq X_b} = 1$ for every $b \in A$ and $\utility{\xi}{\cdot}$ is monotone, it follows that $\utility{\xi}{\sup_{a \in A} X_a} \geq \utility{\xi}{X_b}$. Since $b$ is arbitrary, we have $\utility{\xi}{\sup_{a \in A} X_a} \geq \sup_{a \in A} \utility{\xi}{X_a}$. Next, since $\state{h+1} = \phi(\state{h},\action{h},\noise{h})$, (\ref{eqn:utility_bellman_equation}) and the monotonicity and translation invariance of $\mathcal{U}{}$:
    \begin{align*}
        \utilityopt{0}{\state{0}} 
        &= \sup_{\action{0}}\, \utility{\noise{0}}{\reward{\state{0}}{\action{0}} + \utilityopt{1}{\state{1}}} \\
        &= \sup_{\action{0}}\, \utility{\noise{0}}{\reward{\state{0}}{\action{0}} + \sup_{\action{1}}\, \utility{\noise{1}}{\reward{\state{1}}{\action{1}} + \utilityopt{2}{\state{2}}}} \\
        &\geq \sup_{\action{0}}\, \sup_{\action{1}}\, \utility{\noise{0}}{\reward{\state{0}}{\action{0}} + \utility{\noise{1}}{\reward{\state{1}}{\action{1}} + \utilityopt{2}{\state{2}}}} \\
        &= \sup_{\action{0},\action{1}}\, \utility{\noise{0}}{\utility{\noise{1}}{\reward{\state{0}}{\action{0}} + \reward{\state{1}}{\action{1}} + \utilityopt{2}{\state{2}}}} \\
        &= \sup_{\action{0},\action{1}}\, \utility{\noise{0},\noise{1}}{\reward{\state{0}}{\action{0}} + \reward{\state{1}}{\action{1}} + \utilityopt{2}{\state{2}}} \\
        &\geq \dots \\
        &= \sup_{\action{0:H}}\, \utility{\noise{0:H}}{\reward{\state{0}}{\action{0}} + \dots + \reward{\state{H}}{\action{H}}} \\
        &= u_{SL}(\state{0}).
    \end{align*}
    This completes the proof.

\subsection{Additional Experiments}
\label{sec:ablation}

\subsubsection{Varying Risk-Aversion}
\label{ssec:varying-beta}
\begin{figure}[h!]
    \centering
    \begin{subfigure}[c]{0.33\textwidth}
        \centering
           \includegraphics[width=0.99\linewidth]{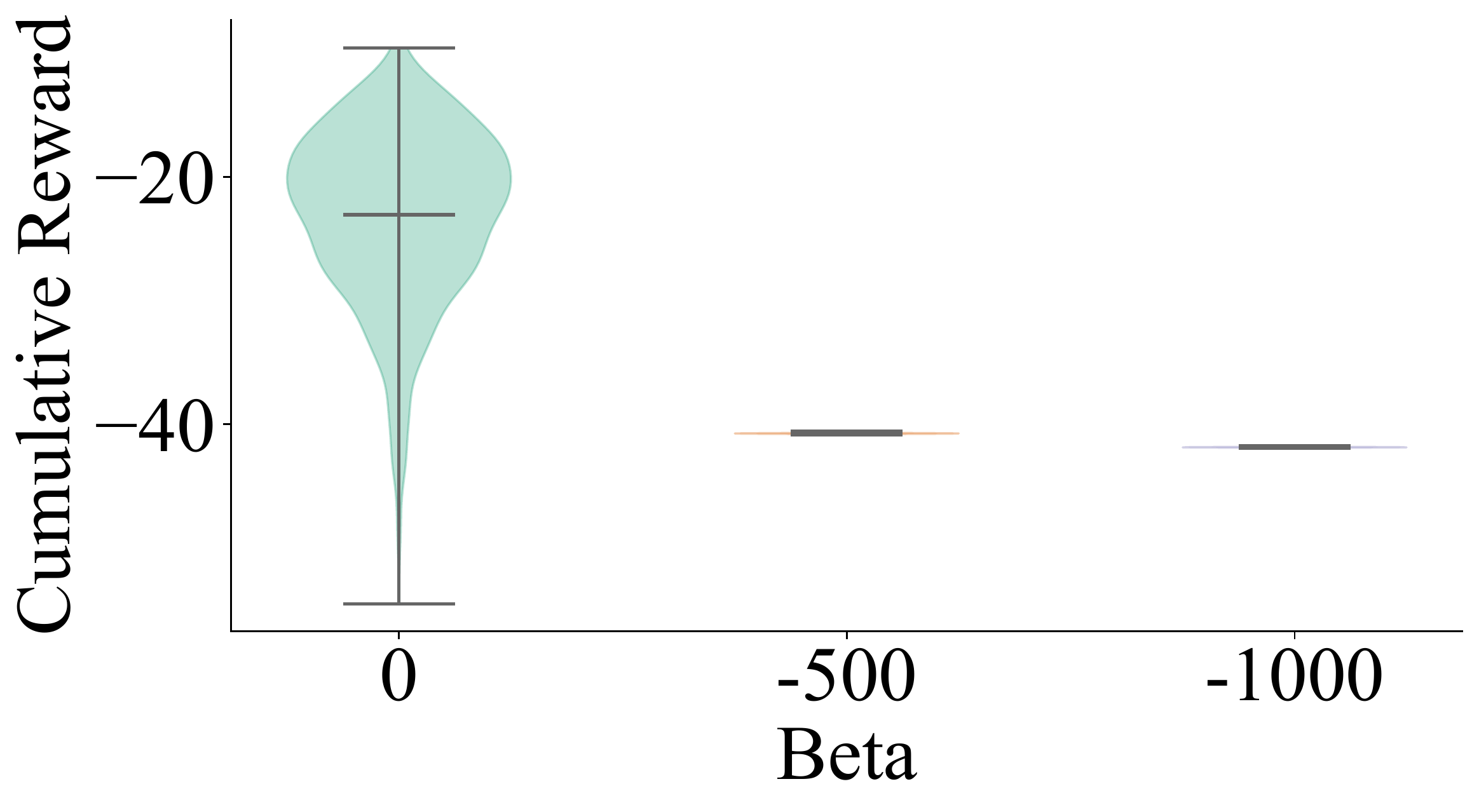}
        \caption{Navigation}
    \end{subfigure}%
    \begin{subfigure}[c]{0.33\textwidth}
        \centering
        \includegraphics[width=0.99\linewidth]{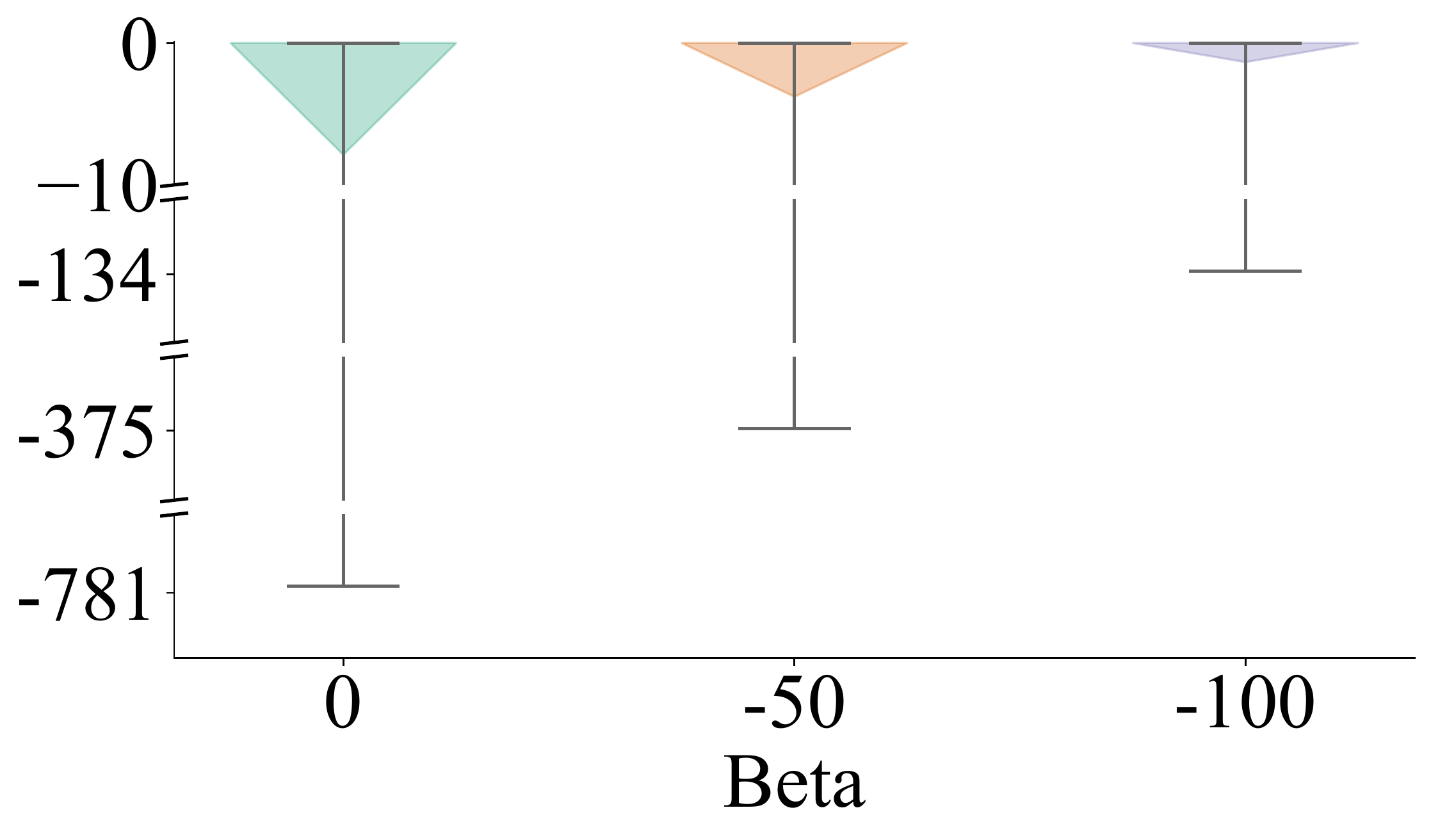}
        \caption{Reservoir}
    \end{subfigure}%
    \begin{subfigure}[c]{0.33\textwidth}
        \centering
           \includegraphics[width=0.99\linewidth]{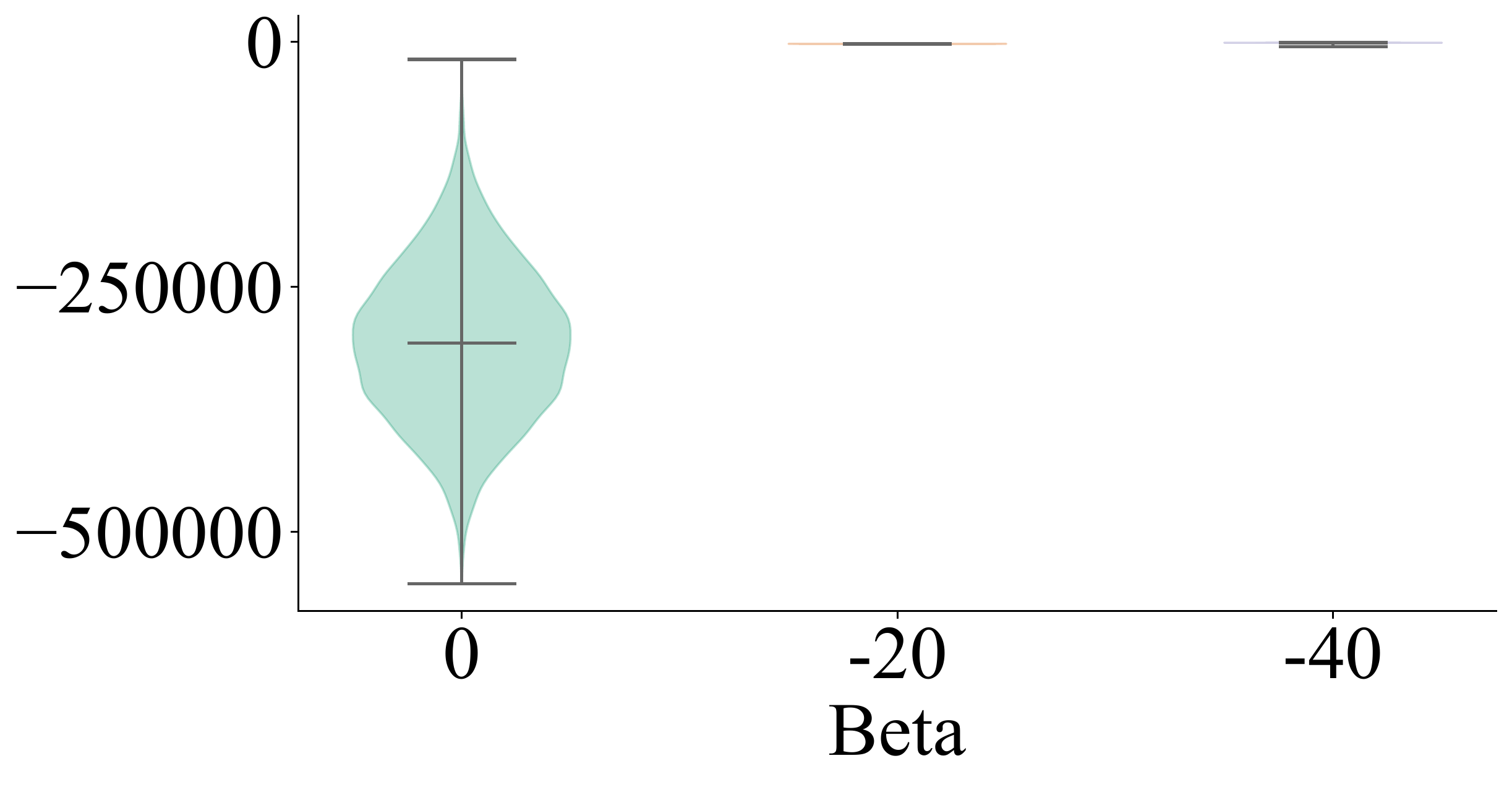}
        \caption{HVAC}
    \end{subfigure}%
    \caption{The cumulative reward distributions for \acronym-DRP over different values of the risk-aversion parameter $\beta$. Risk aversion increases from left to right.}
    \label{fig:triple_reward_dists}
\end{figure}

In this section, we evaluate the performance of \acronym{}-DRP with varying values of the risk-aversion parameter $\beta$ in order to investigate its impact. Three levels of risk-averseness are chosen per each domain: risk-neutral, intermediate risk-aversion, and high risk-aversion. As in Section \ref{sec:experiments}, we use the mean-variance approximation of the entropic risk \eqref{eqn:symbolic_mv} for all evaluations. Here, smaller $\beta$ (i.e., negative $\beta$ with large magnitude) represents greater risk-averseness.

In Figure \ref{fig:triple_reward_dists}, we can see that the $\beta$ parameter indeed gives us the control over how much risk-aversion we want an optimized policy to have. For Navigation and HVAC, we could obtain great reductions in the variances of cumulative rewards even with the intermediate $\beta$ values. In Reservoir, it is more apparent that \acronym{}-DRP with smaller $\beta$ values show more risk-averse behavior as represented by less variability in the corresponding returns. In general, the tradeoff is that we would attain a higher expected return yet a greater variance in returns with a higher $\beta$. However, DRP ($\beta=0$) gets lower expected returns in Reservoir and HVAC. As discussed in Section \ref{ssec:discussion}, this is because large penalties are incurred when visiting states that are outside of some bounds in these domains. 
\subsubsection{The Exact Entropic Risk Measure}

In addition, we evaluate SLP, DRP, \acronym{}-SLP, and \acronym{}-DRP in all three domains using the exact entropic risk measure \eqref{eqn:entropic}. Due to the well-known numerical instability of computation using the exact entropic risk \cite{gosavi2014beyond}, we are limited to $\beta$ with small magnitudes in these experiments: $\beta=-10^{-3}$ for Navigation and Reservoir, and $\beta = -10^{-4}$ for HVAC. 

In Figure \ref{fig:entropic_cumulative_reward_dists}, we find similar patterns exhibited in Figure \ref{fig:cumulative_reward_dists}. That is, DRP performs better than SLP, and \acronym{} agents show smaller variance in returns than their risk-neutral counterparts. However, since we could only use the $\beta$ values with small magnitudes due to the aforementioned numerical instability, the reductions in variances are not significant when compared to the results in Figure \ref{fig:cumulative_reward_dists}. On the other hand, the results also indicate that it is possible to obtain comparable (for Navigation and Reservoir) or higher (for HVAC) expected returns with \acronym{} methods. When combined with the analysis in Appendix \ref{ssec:varying-beta}, these results suggest that it is possible to find $\beta$ that is sufficiently small such that the expected return is not excessively penalized but is large enough to reduce the variance of returns. In conclusion, we note that higher-levels of risk-aversion would necessitate using the mean-variance approximation (as experimented in the main paper) due to its numerical stability across a wider range of $\beta$ values.


\begin{figure}[h!]
    \centering
    \begin{subfigure}[c]{0.33\textwidth}
        \centering
           \includegraphics[width=0.99\linewidth]{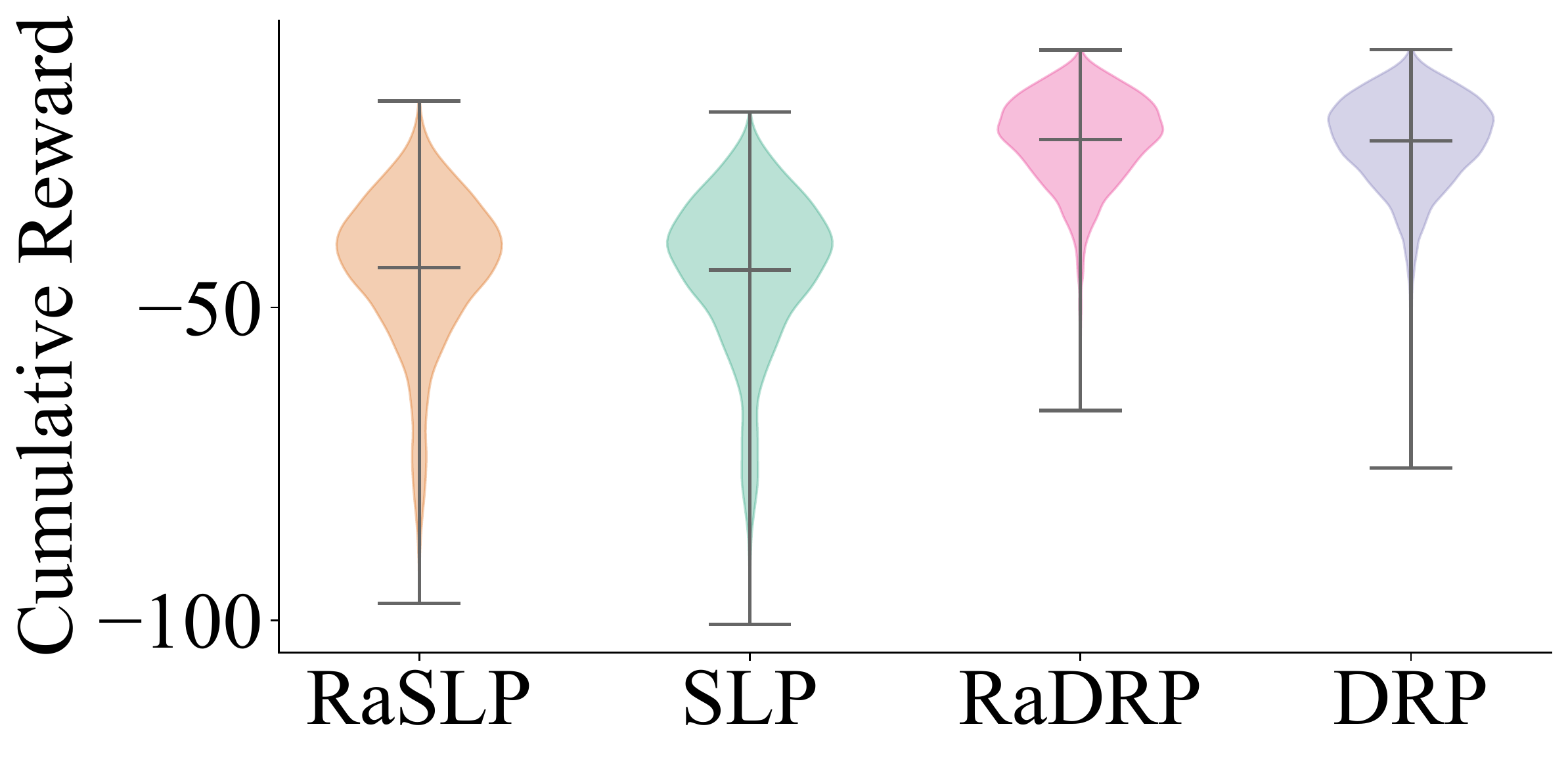}
        \caption{Navigation}
    \end{subfigure}%
    \begin{subfigure}[c]{0.33\textwidth}
        \centering
        \includegraphics[width=0.99\linewidth]{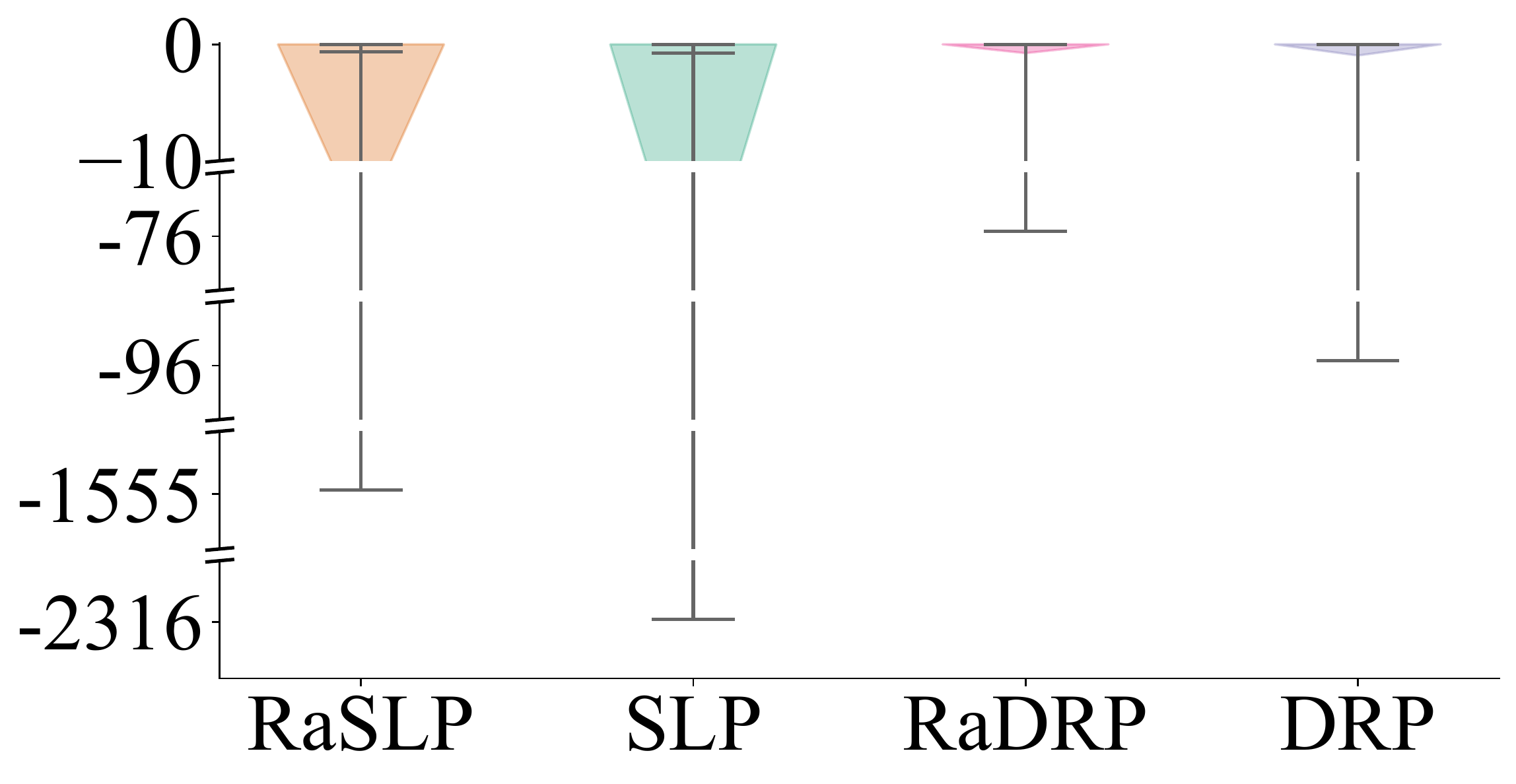}
        \caption{Reservoir}
    \end{subfigure}%
    \begin{subfigure}[c]{0.33\textwidth}
        \centering
           \includegraphics[width=0.99\linewidth]{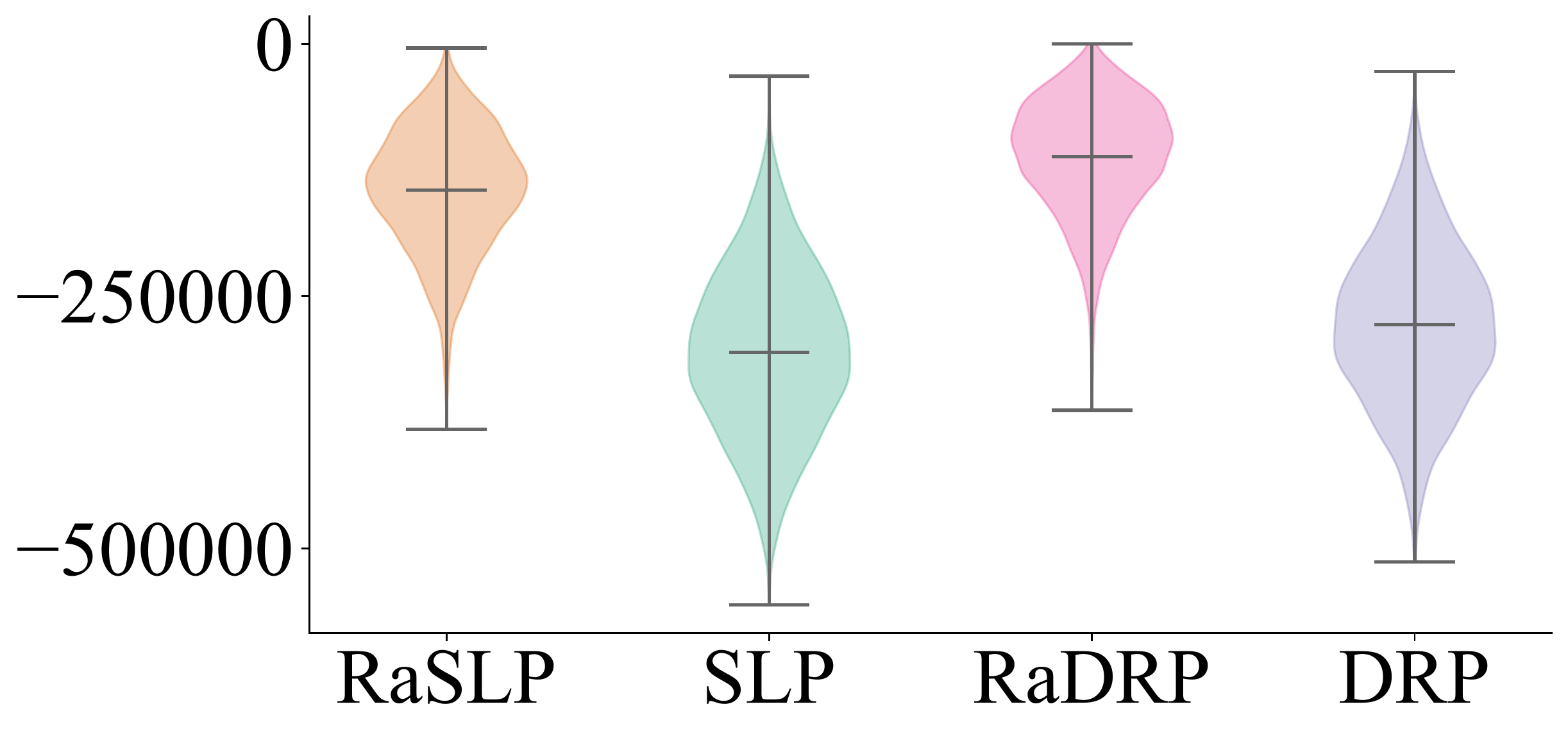}
        \caption{HVAC}
    \end{subfigure}%
    \caption{The cumulative reward distributions for SLP, \acronym-SLP, DRP and \acronym-DRP in all three domains using the exact entropic risk measure instead of the mean-variance approximation.}
    \label{fig:entropic_cumulative_reward_dists}
\end{figure}

\subsection{Experimental Setting Details}
\label{ssec:detailed_experiment}

In this section, we provide a detailed accounting of the experimental parameters used in all experiments.  All experiments were run on a single compute node with AMD Ryzen 7 4800H processor and NVIDIA GeForce RTX 2060 graphics card using PyTorch v1.7.1 and Python version 3.8.5. On this machine, compute time for each training run of a domain took less than 3 minutes. 
$\beta$ was chosen by iteratively increasing $\beta$ from $0$ until a significant reduction in variance was achieved for \acronym-SLP. The same $\beta$ was then used for \acronym-DRP to make for fair comparision. 

\newcommand\Tstrut{\rule{0pt}{2.6ex}}         
\newcommand\Bstrut{\rule[-0.9ex]{0pt}{0pt}}   

\begin{table}[h!]
    \centering
    \begin{tabular}{c|c|c|c|c}
        \toprule
        Parameter & SLP & \acronym-SLP & DRP & \acronym-DRP
        \Bstrut\\
        \hline\hline
        \Tstrut
        Learning Rate & $0.5$ & $0.5$ & $2.5 * 10^{-4}$ & $2.5 * 10^{-4}$\\
        
        Training Epochs & $1001$ & $1001$ & $1001$ & $1001$ \\
        
        Batch Size & $8192$ & $8192$ & $8192$ & $8192$ \\
        
        $\beta$ & $0$ & $-1000$ & $0$ & $-1000$ \\
        
        Time Horizon & $20$ & $20$ & $20$ & $20$ \\
        
        Hidden-Layer-Nodes & N/A & N/A & $256,128,64,32$ & $256,128,64,32$ \\ \\
    \end{tabular}
    \caption{Hyperparameters of trained agents in the Navigation domain.}
    \label{tab:nav_hyperparameter_breakdown}
\end{table}

\begin{table}[h!]
    \centering
    \begin{tabular}{c|c|c|c|c}
        \toprule
        Parameter & SLP & \acronym-SLP & DRP & \acronym-DRP
        \Bstrut\\
        \hline\hline
        \Tstrut
        Learning Rate & $0.2$ & $0.2$ & $5 * 10^{-3}$ & $5 * 10^{-3}$\\
        
        Training Epochs & $501$ & $501$ & $501$ & $501$ \\
        
        Batch Size & $1024$ & $1024$ & $1024$ & $1024$ \\
        
        $\beta$ & $0$ & $-100$ & $0$ & $-100$ \\
        
        Time Horizon & $50$ & $50$ & $50$ & $50$ \\
        
        Hidden-Layer-Nodes & N/A & N/A & $256,128,64,32$ & $256,128,64,32$ \\ \\
    \end{tabular}
    \caption{Hyperparameters of trained agents in the Reservoir domain.}
    \label{tab:reservoir_hyperparameter_breakdown}
\end{table}

\begin{table}[h!]
    \centering
    \begin{tabular}{c|c|c|c|c}
        \toprule
        Parameter & SLP & \acronym-SLP & DRP & \acronym-DRP
        \Bstrut\\
        \hline\hline
        \Tstrut
        Learning Rate & $5 * 10^{-3}$ & $5 * 10^{-3}$ & $5 * 10^{-3}$ & $5 * 10^{-3}$\\
        
        Training Epochs & $501$ & $501$ & $501$ & $501$ \\
        
        Batch Size & $128$ & $128$ & $128$ & $128$ \\
        
        $\beta$ & $0$ & $-40$ & $0$ & $-40$ \\
        
        Time Horizon & $125$ & $125$ & $125$ & $125$ \\
        
        Hidden-Layer-Nodes & N/A & N/A & $256,128,64,32$ & $256,128,64,32$ \\ \\
    \end{tabular}
    \caption{Hyperparameters of trained agents in the HVAC domain.}
    \label{tab:hvac_hyperparameter_breakdown}
\end{table}
\fi
\end{document}

%% file: symbols.tex
\usepackage{amsmath}
\usepackage{amsfonts}
\usepackage{amsthm}
\usepackage{amssymb}
\usepackage{mathtools}

\newtheorem{theorem}{Theorem}

\theoremstyle{definition}

\newcommand{\reals}{\mathbb{R}}

\newcommand{\defeq}{\vcentcolon=}

\newcommand{\E}[2]{\mathbb{E}_{#1}\!\left[#2\right]}
\newcommand{\Egiven}[3]{\mathbb{E}_{#1}\!\left[\left.\kern-\nulldelimiterspace#2\,\right|\,#3\right]}
\newcommand{\var}[2]{\mathrm{Var}_{#1}\!\left[#2\right]}
\newcommand{\prob}[1]{\mathbb{P}\!\left(#1\right)}

\newcommand{\states}{\mathcal{S}}
\newcommand{\actions}{\mathcal{A}}
\newcommand{\state}[1]{\mathbf{s}_{#1}}
\newcommand{\action}[1]{\mathbf{a}_{#1}}

\newcommand{\rewardfunction}{r}
\newcommand{\reward}[2]{\rewardfunction(#1,#2)}

\newcommand{\dynamicsfunction}{p}
\newcommand{\dynamics}[3]{\dynamicsfunction\!\left(\left.\kern-\nulldelimiterspace#3 \,\right|\, #1, #2\right)}

\newcommand{\noises}{\Xi}
\newcommand{\noisedensity}{f_{\xi}}
\newcommand{\noise}[1]{\xi_{#1}}

\newcommand{\utilityfunction}{\mathcal{U}}
\newcommand{\utility}[2]{\utilityfunction_{#1}\!\left(#2\right)}
\newcommand{\utilityopt}[2]{{U}^*_{#1}\!\left(#2\right)
}

\newcommand{\pderiv}[2]{\frac{\partial #1}{\partial #2}}